\documentclass[10pt]{article}

\usepackage{amsmath}
\usepackage{amssymb}
\usepackage{amsmath}
\usepackage{graphicx}
\usepackage{cite}
\usepackage{amsthm}
\usepackage{float}
\usepackage{bm}
\usepackage{color,soul}
\usepackage[noend]{algpseudocode}
\usepackage{algorithm,algorithmicx}
\usepackage{lipsum}
\usepackage{subcaption}
\usepackage{url}
\usepackage{mathtools}
\usepackage{titlesec}
\usepackage{fancyhdr}
\usepackage{lastpage}
\usepackage{fullpage}

\usepackage[T1]{fontenc}
\usepackage{textcomp}
\usepackage[charter]{mathdesign}
\usepackage{oldgerm}

\titleformat*{\section}{\large\bfseries}
\titleformat*{\subsection}{\normalsize\bfseries}

\def\defn{\,\triangleq\,}
\def\im{{\mathrm{j}}}
\def\e{{\mathrm{e}}}
\def\dprod{\bullet}

\def\uin{{u_{\text{\tiny in}}}}
\def\usc{{u_{\text{\tiny sc}}}}

\def\ebf{{\mathbf{e}}}
\def\fbf{{\mathbf{f}}}
\def\gbf{{\mathbf{g}}}

\def\pbf{{\mathbf{p}}}
\def\qbf{{\mathbf{q}}}
\def\rbf{{\mathbf{r}}}
\def\sbf{{\mathbf{s}}}
\def\ubf{{\mathbf{u}}}

\def\ybf{{\mathbf{y}}}
\def\zbf{{\mathbf{z}}}

\def\ubfin{{\mathbf{u}_{\text{\tiny in}}}}
\def\ubfinit{{\mathbf{u}_{\text{\tiny init}}}}

\def\gbfhat{{\widehat{\mathbf{g}}}}
\def\ubfhat{{\widehat{\mathbf{u}}}}
\def\fbfhat{{\widehat{\mathbf{f}}}}

\def\fbftilde{{\tilde{\mathbf{f}}}}
\def\gbftilde{{\tilde{\mathbf{g}}}}

\def\Hbf{{\mathbf{H}}}
\def\Dbf{{\mathbf{D}}}

\def\Abf{{\mathbf{A}}}
\def\Ibf{{\mathbf{I}}}
\def\Gbf{{\mathbf{G}}}
\def\Sbf{{\mathbf{S}}}

\def\Dcal{\mathcal{D}}
\def\Fcal{\mathcal{F}}
\def\Gcal{\mathcal{G}}

\def\Qcal{\mathcal{Q}}
\def\Rcal{\mathcal{R}}

\def\Scal{\mathcal{S}}

\def\jrm{\mathrm{j}}
\def\Trm{\text{T}}
\def\Irm{\text{I}}
\def\Hrm{\text{H}}
\def\drm{{\, \mathrm{d}}} 
\def\diag{{\mathrm{diag}}}
\newcommand{\sub}[1]{\text{\tiny #1}}

\def\xbm{{\bm{x}}} 
\def\xbmp{{\bm{x}^\prime}} 

\def\R{\mathbb{R}}
\def\C{\mathbb{C}}

\def\argmin{\mathop{\mathrm{arg\,min}}}
\def\prox{\mathrm{prox}}
\def\proj{\mathrm{proj}}
\def\grad{\nabla}

\newcommand*\Let[2]{\State #1 $\gets$ #2}

\theoremstyle{definition}

\begin{document}


\title{SEAGLE: Sparsity-Driven Image Reconstruction \\under Multiple Scattering}


\author{Hsiou-Yuan~Liu%
\thanks{H.-Y.~Liu (email:~hyliu@eecs.berkeley.edu) and L.~Waller (email:~waller@berkeley.edu) are with the
Department of Electrical Engineering \& Computer Sciences, University of California, Berkeley, CA 94720, USA. This work
was completed while H.-Y.~Liu was with Mitsubishi Electric Research Laboratories (MERL).}
\hspace{0.05em},
Dehong~Liu%
\thanks{D.~Liu (email:~liudh@merl.com), H.~Mansour (email:~mansour@merl.com), P.~T.~Boufounos (email:~petrosb@merl.com), and U.~S.~Kamilov (email:~kamilov@merl.com)
are with Mitsubishi Electric Research Laboratories (MERL), 201 Broadway, Cambridge,
MA 02139, USA.}
\hspace{0.05em}, Hassan~Mansour$^\dagger$,
\\Petros~T.~Boufounos$^\dagger$, Laura~Waller$^\ast$, and Ulugbek~S.~Kamilov$^\dagger$}

\maketitle


\begin{abstract}
Multiple scattering of an electromagnetic wave as it passes through an object is a fundamental problem that limits the performance of current imaging systems. In this paper, we describe a new technique---called Series Expansion with Accelerated Gradient Descent on Lippmann-Schwinger Equation (SEAGLE)---for robust imaging under multiple scattering based on a combination of a new nonlinear forward model and a total variation (TV) regularizer. The proposed forward model can account for multiple scattering, which makes it advantageous in applications where linear models are inaccurate. Specifically, it corresponds to a series expansion of the scattered wave with an accelerated-gradient method. This expansion guarantees the convergence even for strongly scattering objects. One of our key insights is that it is possible to obtain an explicit formula for computing the gradient of our nonlinear forward model with respect to the unknown object, thus enabling fast image reconstruction with the state-of-the-art fast iterative shrinkage/thresholding algorithm (FISTA). The proposed method is validated on both simulated and experimentally measured data.
\end{abstract}


\section{Introduction}
\label{Sec:Introduction}

Reconstruction of the spatial permittivity distribution of an unknown object from the measurements of the scattered wave is common in numerous applications. Traditional formulations of the problem are based on linearizing the relationship between the permittivity and the measured wave. For example, if one assumes a straight-ray propagation of waves, the phase of the transmitted wave can be interpreted as a line integral of the permittivity along the propagation direction. This approximation leads to an efficient reconstruction with the filtered back-projection algorithm~\cite{Kak.Slaney1988}. Diffraction tomography is a more refined linear scattering model based on the first Born or Rytov approximations~\cite{Wolf1969, Devaney1981, Born.Wolf1999}. It establishes a Fourier transform-based relationship between the measured wave and the permittivity, and thus enables the reconstruction of the latter with a single numerical application of the inverse Fourier transform.

Recent research in compressive sensing and sparse signal processing has established that sparse regularization can dramatically improve the quality of reconstructed images, even when the amount of measured data is severely limited~\cite{Candes.etal2006, Donoho2006}. This has popularized optimization-based inverse scattering approaches that combine linear forward models with regularizers that mitigate ill-posedness by promoting solutions that are sparse in a suitable transform domain. One class of such regularizers is total variation (TV)~\cite{Rudin.etal1992}, which was demonstrated to improve imaging quality by substantially reducing undesired artifacts due to missing data~\cite{Bronstein.etal2002, Lim.etal2015, Sung.Dasari2011}.

\begin{figure}[t]
\begin{center}
\includegraphics[width=7.5cm]{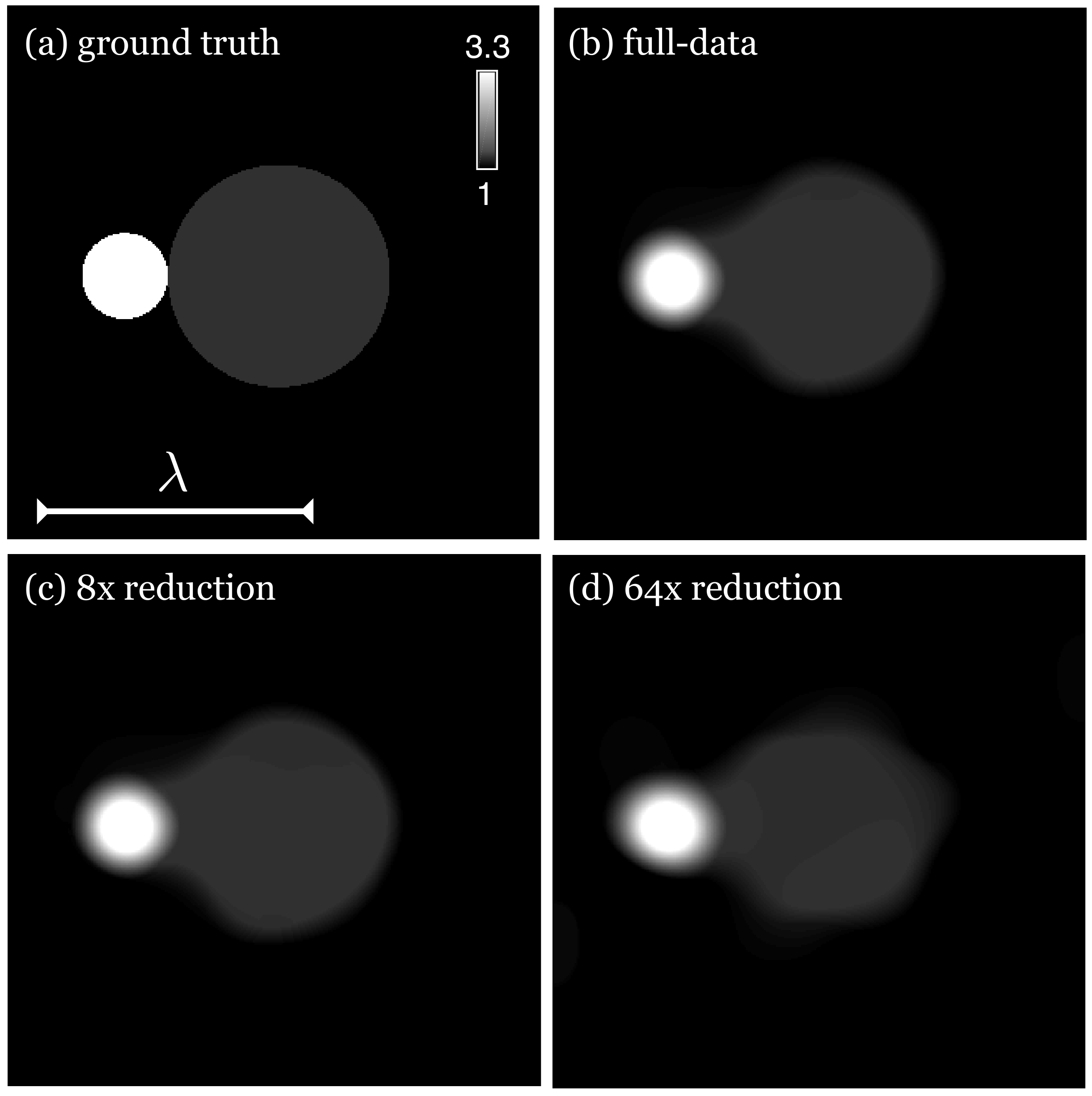}
\end{center}
\caption{The method proposed in this paper can be used to reconstruct the spatial distribution of dielectric permittivity from limited measurements of scattered waves. Illustration on the experimentally measured data at $3$ GHz: (a) ground truth; (b) using full data; (c) $8 \times$ data reduction; (d) $64 \times$ data reduction. The scale bar is equal to the wavelength $\lambda$.}
\label{Fig:MainFigure}
\end{figure}

The main advantage of imaging with linear forward models is that the reconstruction can be reduced to a convex optimization problem that is relatively simple and efficient to solve~\cite{Boyd.Vandenberghe2004, Nocedal.Wright2006, Bioucas-Dias.Figueiredo2007, Beck.Teboulle2009a}. However, such models are only accurate when scattering is weak, which is true when the object is relatively small or has a low permittivity contrast compared to the background~\cite{Chen.Stamnes1998}. When scattering is strong, multiple scattering of waves as they pass through the object limit the validity and accuracy of linear forward models. Multiple scattering is a fundamental problem in tomographic imaging and its complete resolution would enable imaging through strongly scattering objects such as human tissue~\cite{Ntziachristos2010}. As multiple scattering leads to nonlinear forward models, the challenge is in finding computationally tractable methods that can account for the nonlinearity while also accommodating sparsity-driven priors for compressive imaging. To that end, we propose a new method that efficiently combines a nonlinear forward model with the TV regularizer, thus enabling high-quality imaging from a limited number of measurements. Figure~\ref{Fig:MainFigure} provides an example, illustrating the quality of reconstruction for a strongly scattering object with limited data. In particular, the $320 \times 320$ image in Figure~\ref{Fig:MainFigure}(d) was obtained from only $16$ experimentally collected measurements at $3$ GHz from the public dataset~\cite{Geffrin.etal2005}.

\subsection{Contributions}

Our work builds upon prior work on inverse scattering that has been applied to a variety of practical problems in optical, microwave, and radar imaging. The proposed method---called Series Expansion with Accelerated Gradient Descent on Lippmann-Schwinger Equation (SEAGLE)---further extends this work with a new forward model that enables efficient sparsity-driven inversion. The performance of SEAGLE is robust to large permittivity contrasts, data reduction, and measurement noise.

The key contributions of this paper are summarized as follows:

\begin{itemize}

\item A new forward model that iteratively approximates the scattered wave with Nesterov's accelerated-gradient method~\cite{Nesterov1983}. The key benefit is the guaranteed convergence even for objects with large permittivity contrasts.

\item An efficient optimization strategy based on explicit evaluation of the gradient of our forward model with respect to the permittivity. The key benefit of this strategy is that it enables efficient sparsity-driven reconstruction for the nonlinear forward model.

\item An extensive validation of our approach on analytical, simulated, and experimental data. The experimental data used in our evaluations comes from a public dataset~\cite{Geffrin.etal2005}, which enables easy comparisons with several other related approaches.

\end{itemize}

\subsection{Related Work}

The vast majority of current imaging systems---such as optical projection tomography (OPT), diffraction tomography, optical coherence tomography (OCT), digital holography, and subsurface radar---still rely on the linearization of object-wave interaction~\cite{Sharpe.etal2002, Choi.etal2007, Bronstein.etal2002, Lauer2002, Sung.etal2009, Sung.Dasari2011, Kim.etal2014, Lim.etal2015, Ralston.etal2006, Davis.etal2007, Brady.etal2009, Tian.etal2010, Chen.etal2015a, Jol2009, Leigsnering.etal2014a, Liu.etal2016a}.
However, early work in microwave imaging have shown the promise of accounting for the nonlinear nature of scattering~\cite{Tijhuis1989, Wang.Chew1989, Chew.Wang1990, Kleinman.etal1990}. These have been extended by a large number of techniques incorporating the nonlinear nature of scattering. Several recent publications have reviewed these methods~\cite{Haeberle.etal2010, Mudry.etal2012, Jin.etal2017}, which include conjugate gradient method (CGM)~\cite{Chaumet.Belkebir, Belkebir.etal2005}, contrast source inversion method (CSIM)~\cite{Dubois.etal2005}, hybrid method (HM)~\cite{Mudry.etal2012}, beam propagation method (BPM)~\cite{Tian.Waller2015, Kamilov.etal2015, Kamilov.etal2016}, recursive Born method (RBM)~\cite{Kamilov.etal2016a}, and bounded inversion method (BIM)~\cite{Zhang.etal2016}. Some recent works have explored the idea of statistical modeling of scattering for imaging through diffusive or turbid media~\cite{Charnotskii2015, Liu.etal2015a, Singh.etal2017}.

The key innovation in this paper is in the new forward model, which enables tractable optimization of a cost function that incorporates nonlinear scattering and a non-differentiable regularizer such as TV. As corroborated by experiments, this formulation enables fast, stable, and reliable convergence, even when working with limited amount of data. This paper extends our preliminary work~\cite{Liu.etal2017} by including key mathematical derivations, as well as more extensive validation on experimentally collected data. 

The experimental data used in this paper comes from a public dataset~\cite{Geffrin.etal2005}. Several other methods have been tested on this dataset~\cite{Abubakar.etal2005, Baussard2005, Crocco.etal2005, Dubois.etal2005, Feron.etal2005}. This enables qualitative evaluation of the performance of the proposed technique against some other algorithms based on nonlinear forward models. 
While many of the methods tested on the data use multi-frequency measurements, the results in this paper rely on a single frequency. However, the method uses the latest techniques in large-scale optimization with sparse regularization, which enables subsampling and leads to improvements in imaging performance. We expect the performance of the proposed method to improve further if multi-frequency measurements are incorporated.

\subsection{Outline}

Section~\ref{Sec:ForwardModel} describes the problem formulation and our forward model. Our algorithmic strategy for reconstructing the image is described in Section~\ref{Sec:ImageFormationMethod}. Experimental results validating our approach are presented in Section~\ref{Sec:ExperimentalEvaluation}. Finally, Section~\ref{Sec:Conclusion} discusses our findings and concludes. We additionally provide three appendices that disclose technical aspects of the proposed method.


\section{Forward model}
\label{Sec:ForwardModel}

\begin{figure}
\centering
\includegraphics[width=5.5cm]{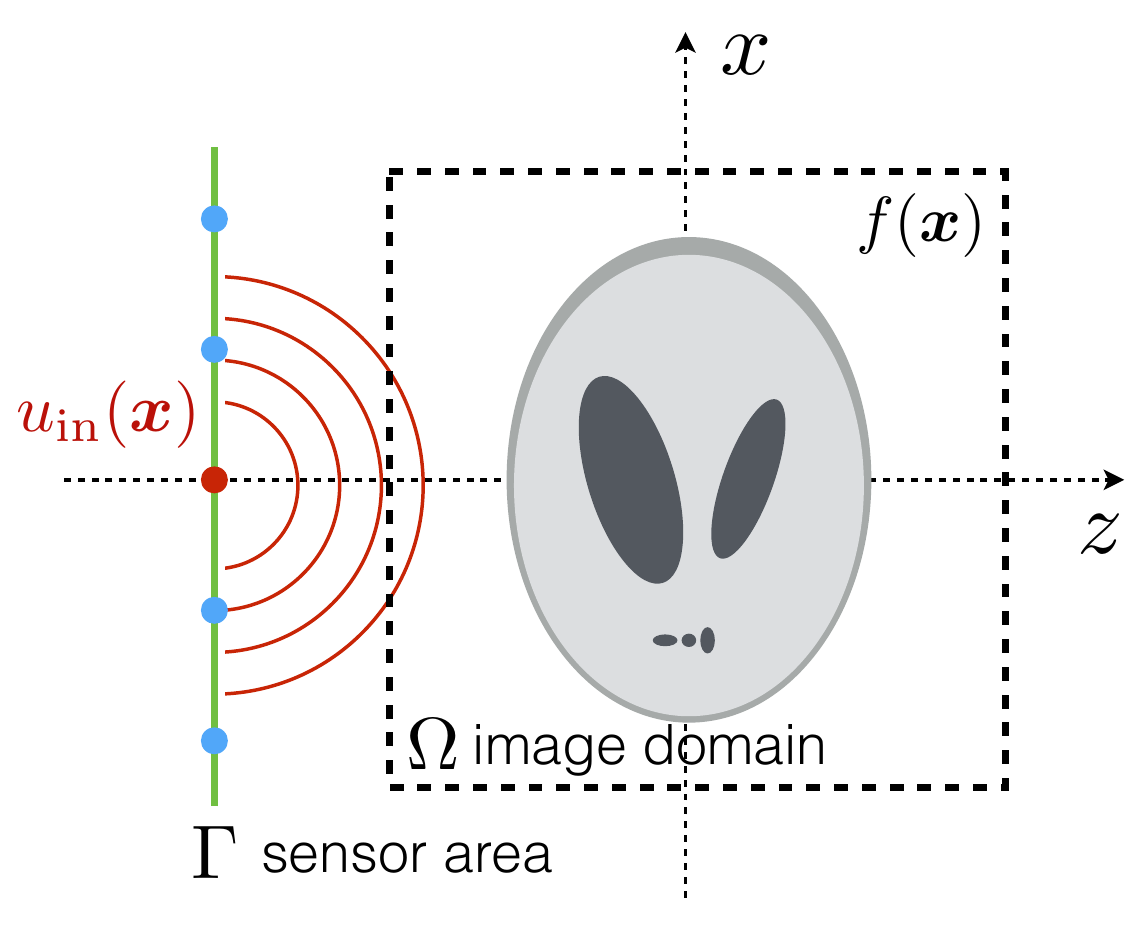}
\caption{\label{Fig:Scheme} Schematic representation of the scattering experiment. An object with a real scattering potential $f(\xbm), \xbm \in \Omega$, is illuminated with an input wave $\uin$, which interacts with the object and leads to the scattered wave $\usc$ measured at the sensing region $\Gamma$ represented with a green line.}
\end{figure}

The forward problem computes the scattered field given a distribution of inhomogeneous permittivity, while the inverse problem reconstructs this distribution. 
The model we propose here can be interpreted as a series expansion based on the iterates of the gradient method. This expansion can be made arbitrarily accurate and is stable for high permittivity objects. Additionally, it enables efficient computation of the gradient of the cost function, which is essential for fast image reconstruction. In this section we focus on introducing the forward model, leaving the gradient evaluation and inversion to the next section. Our derivations are for the scenario of a single illumination, but the generalization to an arbitrary number of illuminations is straightforward.

\subsection{Problem formulation}

Consider the scattering problem in Figure~\ref{Fig:Scheme}, where an object of the permittivity distribution $\epsilon(\xbm)$ in the bounded domain $\Omega \subseteq \R^D$, with $D\in\{2,3\}$, is immersed into a background medium of permittivity $\epsilon_b$, and illuminated with the incident electric field $\uin(\xbm)$. We assume that the incident field is monochromatic and coherent, and it is known inside $\Omega$ and at the locations of the sensors $\Gamma$. The result of object-wave interaction is measured at the location of the sensors as a scattered field $\usc(\xbm)$. The scattering of light can be accurately described by the Lippmann-Schwinger equation inside the image domain~\cite{Born.Wolf2003}
\begin{equation}
\label{Eq:ImageField}
u(\xbm) = \uin(\xbm) + \int_{\Omega} g(\xbm - \xbmp) \, f(\xbmp) \, u(\xbmp) \drm \xbmp,\quad(\xbm \in \Omega)
\end{equation}
where $u(\xbm) = \uin(\xbm) + \usc(\xbm)$ is the total electric field,
${f(\xbm) \defn k^2 (\epsilon(\xbm)-\epsilon_b)}$ is the scattering potential, which is assumed to be real, and $k = 2\pi/\lambda$ is the wavenumber in vacuum. The function $g(\xbm)$ is the Green's function defined as
\begin{equation}
\label{eq:greenfunc}
g(\xbm) \defn 
\begin{dcases}
     \frac{\im}{4} H_0^{(1)}(k_b \|\xbm\|_{\ell_2}) & \text{in 2D}\\
     \frac{\e^{\im k_b \|\xbm\|_{\ell_2}}}{4\pi \|\xbm\|_{\ell_2}} & \text{in 3D},
\end{dcases}
\end{equation}
where $k_b \defn k \sqrt{\epsilon_b}$ is the wavenumber of the background medium and $H_0^{(1)}$ is the zero-order Hankel function of the first kind. Note that the Green's function satisfies the Helmholtz equation 
$$\left(\grad^2+k_b^2 \Irm\right)g(\xbm)=-\delta(\xbm),$$ 
as well as the outgoing-wave boundary condition, and a time-dependence convention under which the physical electric field equals to $\mathrm{Re}\left\{u(\xbm)\e^{-\im\omega t}\right\}$.
The knowledge of the total-field $u$ inside the image domain $\Omega$ enables the prediction of the scattered field at the sensor area
\begin{equation}
\label{Eq:SensorField}
\usc(\xbm) = \int_\Omega g(\xbm-\xbmp) \, f(\xbmp) \, u(\xbmp) \drm \xbmp.\quad(\xbm \in \Gamma)
\end{equation}

The computation of the scattered wave is equivalent to solving~\eqref{Eq:ImageField} for $u(\xbm)$ inside the image and evaluating~\eqref{Eq:SensorField} for $\usc(\xbm)$ at the sensor locations. Note that $u$ is present on both sides of~\eqref{Eq:ImageField} and that the relation between $u$ and the scattering potential $f$ is nonlinear. The first Born and the Rytov approximations~\cite{Wolf1969, Devaney1981, Born.Wolf1999} are linear models that replace $u$ in~\eqref{Eq:SensorField} with a suitable approximation that decouples the nonlinear dependence of $u$ on $f$. However, such linearization imposes a strong assumption that the object is weakly scattering, which makes the corresponding reconstruction methods not applicable to a large variety of imaging problems~\cite{Chen.Stamnes1998}. Our forward model described next is a fast nonlinear method for solving~\eqref{Eq:ImageField} that overcomes the weakly scattering object assumptions.

\subsection{Algorithmic Expansion of the Scattered Waves}

\begin{algorithm}[t]
\caption{Forward model computation\label{alg:forward}}
  \begin{algorithmic}[1]
  \Statex{\textbf{intput: } Image $\fbf \in \R^N$, maximum number of iterations $K$, tolerance $\delta_{\text{\tiny tol}}$, and initialization $\ubfinit =\ubfin$.}
        \Statex\textbf{set: }  $\ubf^{-1} \gets \ubfinit,\ \ubf^0 \gets \ubfinit,\ t_0 \gets 0$
        \For{$k \gets 1 \textrm{ to } K$}
            \State $t_k \gets (1+\sqrt{1+4\smash[b]{t_{k-1}^2}})/{2}$, 
            \State $\mu_k \gets (1-{t_{k-1}})/{t_k}$
            \Let{$\sbf^k$}{$(1-\mu_k)\ubf^{k-1} + \mu_k\ubf^{k-2}$}
            \Let{$\gbf$}{$\Abf^\Hrm(\Abf\sbf^k-\ubf_\sub{in})$} \label{ln:gradient}\Comment{gradient at $\sbf^k$}
            \If{$\| \gbf \|_2 < \delta_{\text{\tiny tol}}$} $K \gets k$, break the loop
            \EndIf
            \Let{$\gamma_k$}{$\smash[b]{\| \gbf \|^2_2 / \| \Abf\gbf \|^2_2}$} \label{ln:forward-lipschitz}
            \Let{$\ubf^k$}{$\sbf^k-\gamma_k\gbf$}  \label{ln:update-u}
        \EndFor
        \Let{$\ubfhat$}{$\ubf^K$}
        \Let{$\zbf$}{$\mathbf{H}(\ubfhat \dprod \fbf)$} \label{ln:last-scat}
        \Statex{\textbf{return:} predicted scattered wave $\zbf$, as well as $\ubfhat$, $\{\sbf^k\}$, $\{\gamma_k\}$, and $\{\mu_k\}$.}
  \end{algorithmic}
\end{algorithm}

We separate the computation of the electric field into two parts: the total field $u(\xbm)$ in the image domain and the scattered field $\usc(\xbm)$ at the sensors. The discretization and combination of~\eqref{Eq:ImageField} and~\eqref{Eq:SensorField} leads to the following matrix-vector description of the forward problem
\begin{subequations}
\label{Eq:NonlinearModel}
\begin{align}
&\ybf = \Hbf(\ubf \dprod \fbf) + \ebf \label{Eq:ForwardScat1}\\
&\ubf = \ubfin + \Gbf (\ubf \dprod \fbf), \label{Eq:ForwardScat2}
\end{align}
\end{subequations}
where $\fbf \in \R^N$ is the discretized scattering potential $f$, $\ybf \in \C^M$ is the measured scattered field $\usc$ at $\Gamma$, $\ubfin \in \C^N$ is the input field $\uin$ inside $\Omega$, $\Hbf \in \C^{M \times N}$ is the discretization of the Green's function at $\Gamma$, $\Gbf \in \C^{N \times N}$ is the discretization of the Green's function inside $\Omega$, the symbol $\bullet$ denotes a component-wise multiplication between two vectors, and $\ebf \in \C^M$ models the random noise at the measurements. Using the shorthand notation $\Abf \defn \Ibf - \Gbf\diag\{\fbf\}$, where $\Ibf \in \R^{N \times N}$ is the identity matrix and $\diag\{\cdot\}$ is an operator that forms a diagonal matrix from its argument, we can represent the forward scattering in~\eqref{Eq:ForwardScat2} as a minimization problem
\begin{align}
\label{Eq:ForwardModel}
&\ubfhat(\fbf) \defn \argmin_{\ubf \in \C^N} \left\{\Scal(\ubf)\right\} \\
&\quad \quad\text{with}\quad \Scal(\ubf) \defn \frac{1}{2}\|\Abf \ubf -\ubfin\|_{\ell_2}^2, \nonumber
\end{align}
where the matrix $\Abf$ is a function of $\fbf$. The gradient of $\Scal$ can be computed as
\begin{equation}
\nabla \Scal(\ubf) = \Abf^\Hrm(\Abf\ubf - \ubfin).
\end{equation}
Since~\eqref{Eq:ForwardModel} corresponds to the optimization of a differentiable function, it is possible to compute the total field $\ubfhat$ iteratively using Nesterov's accelerated-gradient method~\cite{Nesterov1983}
\begin{subequations}
\label{Eq:Nesterov}
\begin{align}
\label{Eq:Nesterov1} &t_k \leftarrow \frac{1}{2}\left(1+\sqrt{1 + t_{k-1}^2}\right) \\
\label{Eq:Nesterov2} &\sbf^k \leftarrow \ubf^{k-1} + ((t_{k-1}-1)/t_k)(\ubf^{k-1}-\ubf^{k-2})\\
\label{Eq:Nesterov3} &\ubf^k \leftarrow \sbf^k - \nu \Abf^\Hrm(\Abf\sbf^k - \ubfin),
\end{align}
\end{subequations}
for $k = 1, 2, \dots, K$, where $\ubf^{0} = \ubf^{-1} = \ubfin$, $q_0 = 1$, and $\nu > 0$ is the step-size. At any moment, the predicted scattered field can be set to $\zbf^k = \Hbf(\ubf^k \dprod \fbf)$ with $\ubf^k$ given by~\eqref{Eq:Nesterov3}. Note that the resulting set of fields $\{\ubf^k\}_{k \in [1 \dots K]}$ and $\{\zbf^k\}_{k \in [1 \dots K]}$ can be interpreted as a $K$-term series expansion of the wave-fields inside the object and at the sensor locations, respectively. The full procedure for forward computation with a convenient adaptive step-size is summarized in Algorithm \ref{alg:forward}. 


\section{Inverse Problem}
\label{Sec:ImageFormationMethod}

\begin{figure*}
\centering
\includegraphics[width=13cm]{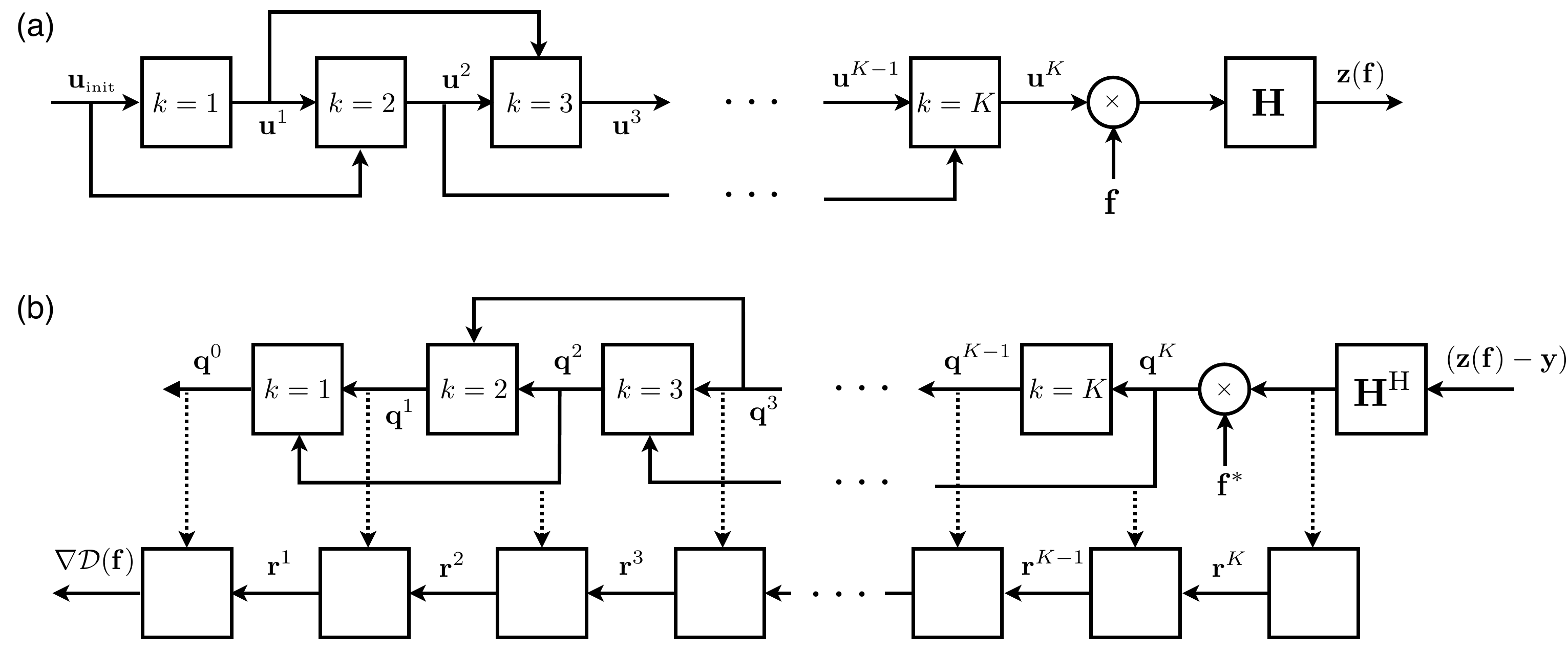}
\caption{\label{Fig:Diagram} Schematic representation of (a) forward model computation in Algorithm~\ref{alg:forward} and (b) error backpropagation in Algorithm~\ref{alg:backward}. Each square box represents basic operations performed for the updates in that iteration.}
\end{figure*}

We now present the overall image reconstruction algorithm, based on the state-of-the-art fast iterative shrinkage/thresholding algorithm (FISTA)~\cite{Beck.Teboulle2009}. The application of FISTA to nonlinear inverse scattering is, however, nontrivial due to the requirement of the gradient of the scattered field with respect to the object. We solve this by providing an explicit formula, based on the error backpropagation method~\cite{Bishop1995}, for computing this gradient.

\subsection{Image Reconstruction}

\begin{algorithm}[t]
\caption{Image formation with FISTA}
\label{Algo:FISTA}
\begin{algorithmic}[1]
\Statex{\textbf{input: }scattered field $\ybf$, initial guess $\fbf^0$, step $\gamma > 0$, and regularization parameter $\tau > 0$.}
\Statex{\textbf{set: } $t \leftarrow 1$, $\fbftilde^0 \leftarrow \fbf^0$, $q_0  \leftarrow 1$}
	\Repeat
	\State $\fbf^t  \leftarrow \prox_{\gamma \tau \Rcal}(\fbftilde^{t-1} - \gamma \nabla\Dcal(\fbftilde^{t-1}))$
	\State $q_t  \leftarrow \frac{1}{2}\left(1 + \sqrt{1+4q_{t-1}^2}\right)$
	\State $\fbftilde^{t}  \leftarrow \fbf^t + ((q_{t-1}-1)/q_t)(\fbf^t - \fbf^{t-1})$
	\State $t \leftarrow t+1$
	\Until{stopping criterion}
	\Statex{\textbf{return:} estimate of the scattering potential $\fbf^t$.}
\end{algorithmic}
\end{algorithm}

\begin{algorithm}[t]
\caption{Error backpropagation for $\nabla \Dcal(\fbf)$\label{alg:backward}}
  \begin{algorithmic}[1]
    \Statex{\textbf{intput: }Image $\fbf \in \R^N$, measurements $\ybf \in \C^M$, input wave field $\ubfin \in \C^N$.}
	\State $(\zbf, \ubfhat, \{\sbf^k\}, \{\gamma_k\}, \{\mu_k\}) \leftarrow$ run forward model in Alg.~\ref{alg:forward}
        \State $\qbf^{K+1} \gets \mathbf{0}$
        \Let{$\qbf^K$}{$\diag\{\fbf\}^\Hrm \Hbf^\Hrm (\zbf-\ybf)$}
        \Let{$\rbf^K$}{$\diag\{\ubfhat\}^\Hrm \Hbf^\Hrm (\zbf-\ybf)$}
        \For{$k \gets K \textrm{ to } 1$}
            \State {$\Sbf^k$} $\triangleq$ {$\Ibf-{\gamma_k}\Abf^\Hrm\Abf$}
            \State {$\mathbf{T}^k$} $\triangleq$ {$\diag\{\Gbf^\Hrm (\Abf\sbf^k-\ubf_\sub{in})\}^\Hrm+\diag\{\sbf^k\}^\Hrm \Gbf^\Hrm \Abf$}
            \Let{$\qbf^{k-1}$}{$(1-\mu_k)\Sbf^k\qbf^k+\mu_{k+1}\Sbf^{k+1}\qbf^{k+1}$}
            \Let{$\rbf^{k-1}$}{$\rbf^k+{\gamma_k}\mathbf{T}^k\qbf^k$}
        \EndFor
        \Statex{\textbf{return:} $\grad\Dcal(\fbf) = \mathrm{Re}\{\rbf^0\}$ the gradient in~\eqref{Eq:Gradient}.}
  \end{algorithmic}
\end{algorithm}

We formulate image reconstruction as the following optimization problem
\begin{subequations}
\label{Eq:ImageFormation}
\begin{equation}
\label{Eq:ImageFormation1}
\fbfhat = \argmin_{\fbf \in \R^N} \left\{\Dcal(\fbf) + \Rcal(\fbf)\right\},
\end{equation}
where
\begin{align}
\label{Eq:ImageFormation2}
&\Dcal(\fbf) \defn \frac{1}{2}\|\ybf - \zbf(\fbf)\|_{\ell_2}^2 \quad\text{and}\\ 
\label{Eq:ImageFormation3}
&\Rcal(\fbf) \defn \tau \sum_{n = 1}^N \|[\Dbf \fbf]_n\|_{\ell_2} = \tau \sum_{n = 1}^N \sqrt{\sum_{d = 1}^D |[\Dbf_d \fbf]_n|^2}.
\end{align}
\end{subequations}
The data-fidelity term $\Dcal$ measures the discrepancy between the actual measurements $\ybf$ and the ones predicted by our scattering model $\zbf$. The function $\Rcal$ is the isotropic TV regularizer with the parameter $\tau > 0$ controlling the strength of the regularization, where $\Dbf: \R^N \rightarrow \R^{N \times D}$ is the discrete gradient operator with matrix $\Dbf_d$ denoting the finite difference operation along dimension $d$ (see Appendix~\ref{Sec:TVMinimization} for details).

The image can then be formed iteratively using a first order method such as ISTA~\cite{Figueiredo.Nowak2003, Daubechies.etal2004, Bect.etal2004}
\begin{equation}
\fbf^t \leftarrow \prox_{\gamma \Rcal} \left(\fbf^{t-1} - \gamma \nabla \Dcal(\fbf^{t-1})\right),
\end{equation}
for $t = 1, 2, 3, \dots$ or its accelerated variant FISTA~\cite{Beck.Teboulle2009} summarized in Algorithm~\ref{Algo:FISTA}. Here, $\gamma > 0$ is a step-size. The operator $\prox_{\gamma \Rcal}$ denotes the proximity operator, and for isotropic TV it can be efficiently evaluated with an algorithm described in Appendix~\ref{Sec:TVMinimization}. Finally, an efficient implementation of the imaging algorithm requires the gradient of the data-fidelity term
\begin{equation}
\label{Eq:Gradient}
\nabla \Dcal(\fbf) = \mathrm{Re}\left\{\left[\frac{\bm{\partial} \, \zbf(\fbf)}{\bm{\partial} \fbf}\right]^\Hrm \left(\zbf(\fbf) - \ybf\right)\right\},
\end{equation}
which can be evaluated using the algorithm summarized in Algorithm~\ref{alg:backward}. The mathematical derivation of this algorithm is similar to that of the standard error backpropagation used in deep learning~\cite{Bishop1995, LeCun.etal2015}, and its full derivation for multiple scattering is provided in Appendix~\ref{Sec:EBP}. As illustrated in Figure~\ref{Fig:Diagram}, the proposed algorithm uses the recursive structure of accelerated-gradient expansion in order to obtain an efficient formula for evaluating the gradient of the data-fidelity term.  While the theoretical convergence of FISTA is difficult to analyze for nonconvex functions, it is often used as a faster alternative to the standard gradient-based methods in the context deep learning and broader machine learning~\cite{Bottou2012, Mairal.etal2014, Bottou.etal2016}. In fact, we observed that our method reliably converges and achieves excellent results on a wide array of problems, as reported in Section~\ref{Sec:ExperimentalEvaluation}. 


\section{Experimental Evaluation}
\label{Sec:ExperimentalEvaluation}

\begin{figure}[t]
\centering
\includegraphics[width=8.5cm]{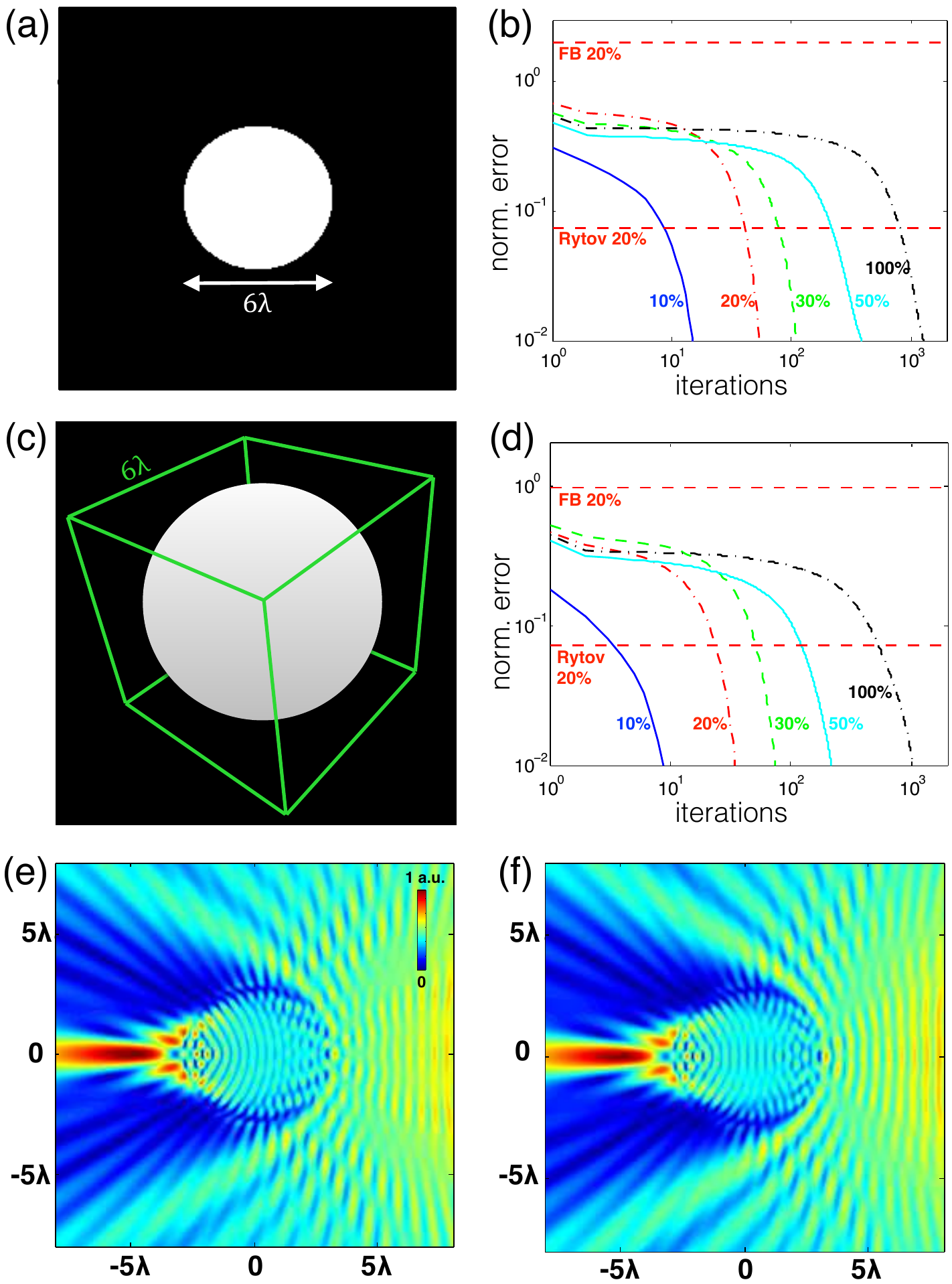}
\caption{\label{fig:compAnalytic}
Analytical validation of the forward model: 
(a) cylinder of diameter $6\lambda$; 
(b) normalized errors in scattering for different contrast levels;
(c) sphere of diameter $6\lambda$; 
(d) normalized errors in scattering for different contrast levels;
(e) analytic field for a cylinder at a contrast level of 100\%;
(f) corresponding field computed by our forward model.}
\end{figure}

\begin{figure}[t]
\centering
\includegraphics[width=8.5cm]{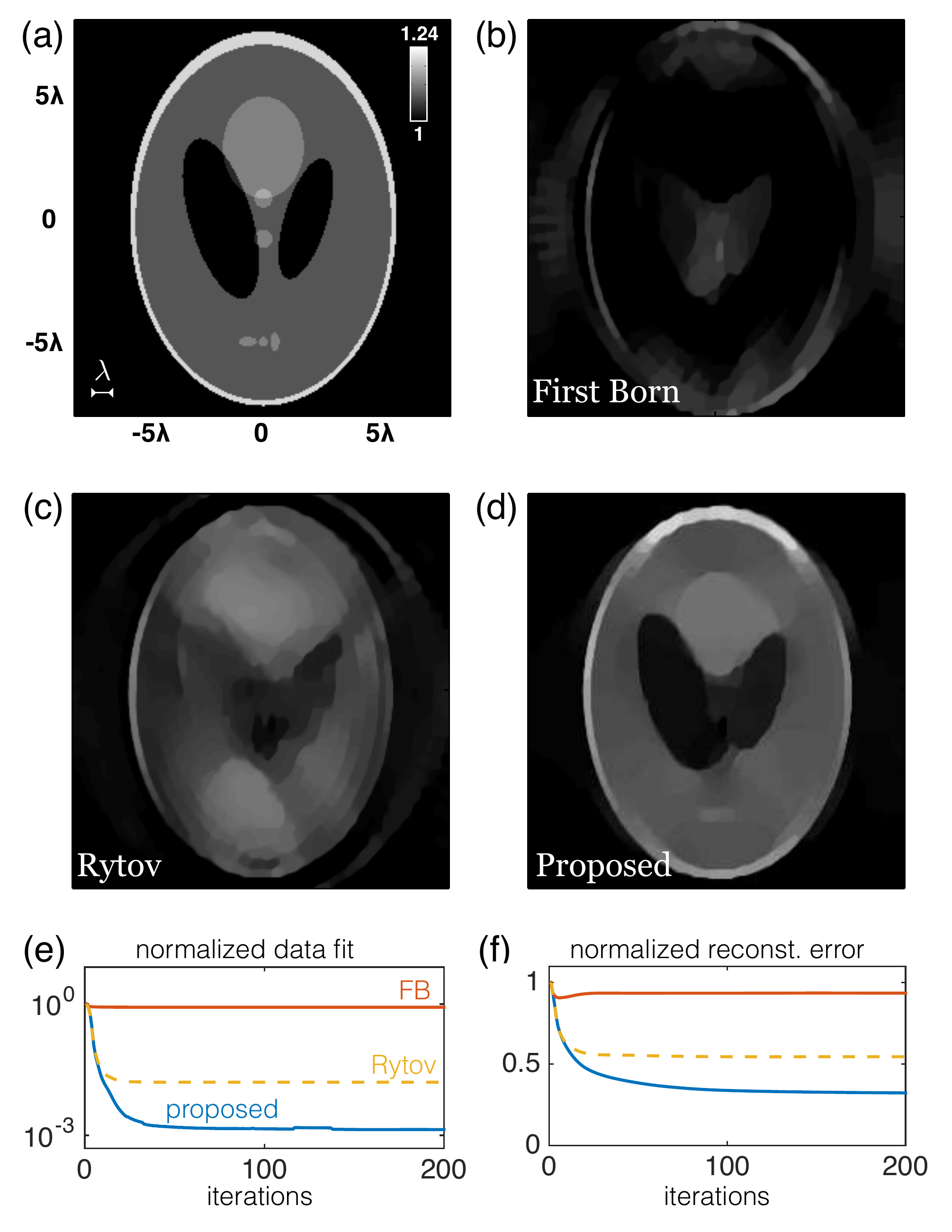}
\caption{\label{fig:inverse} Comparison of the proposed approach with baseline methods on simulated data. 
(a) Shepp-Logan at $20\%$ contrast;
the reconstructed results with (b) the first Born approximation; (c) the Rytov approximation, and (d) our method;
(e) per-iteration data fit;
(f) per-iteration error.}
\end{figure}

We now present the results of validating our method on analytically obtained scattering data for simple scenarios, scattering data obtained with a high-fidelity finite-difference time-domain (FDTD) simulator~\cite{Taflove.Hagness2005}, and experimentally collected data from the public dataset~\cite{Geffrin.Sabouroux2009}.

\subsection{Validation on analytic data}

In the first set of experiments, we validated our forward model for two simple objects where analytic expressions of the scattered wave exist: a two-dimensional point source scattered by a cylinder, and a three-dimensional point source scattered by a sphere. The expressions are derived following the mathematical formalism in~\cite{Jackson1999}, which we review in Appendix~\ref{Sec:Analytic} for completeness. As illustrated in Figures~\ref{fig:compAnalytic}(a) and (b), in both cases, the objects have diameters equal to $6$ wavelengths. The wavelength is set to $74.9$~mm, the source is placed $1$~m away from the center of the objects, the grid size is set to $4.8$~mm ($6$~mm), and there are $250$ points ($128$ points) along each axis in 2D (3D). The contrast of an object is defined as $\max(|\fbf|)/k_b^2$. In Figures~\ref{fig:compAnalytic}(b) and~(d), we quantitatively evaluate the performance of our forward model with the normalized error defined as
\begin{equation}
\text{normalized error}\defn \frac{\|\ubfhat-\ubf_{\text{\tiny true}}\|_{\ell_2}^2}{\|\ubf_{\text{\tiny true}}\|_{\ell_2}^2},
\end{equation}
where $\ubfhat$ is the solution of~\eqref{Eq:ForwardModel} and $\ubf_{\text{\tiny true}}$ is the analytic expression. For comparison, we additionally provide the performance of the first Born (FB) and Rytov approximations at $20\%$ contrast. In Fig.~\ref{fig:compAnalytic}(e) and \ref{fig:compAnalytic}(f), we demonstrate a visual comparison between the analytic expression and the result of our model. Overall, we observed that, by allowing for a large enough value of $K$, our forward model can match the analytically obtained field with arbitrarily high precision. The actual value of $K$ depends on the severity of multiple scattering and must be adapted on the basis of the application of interest. For example, we observed that for objects closely resembling biological samples, one generally requires $10 \leq K \leq 30$.

\subsection{Validation on FDTD simulations}

\begin{figure*}[t]
\centering
\includegraphics[width=15cm]{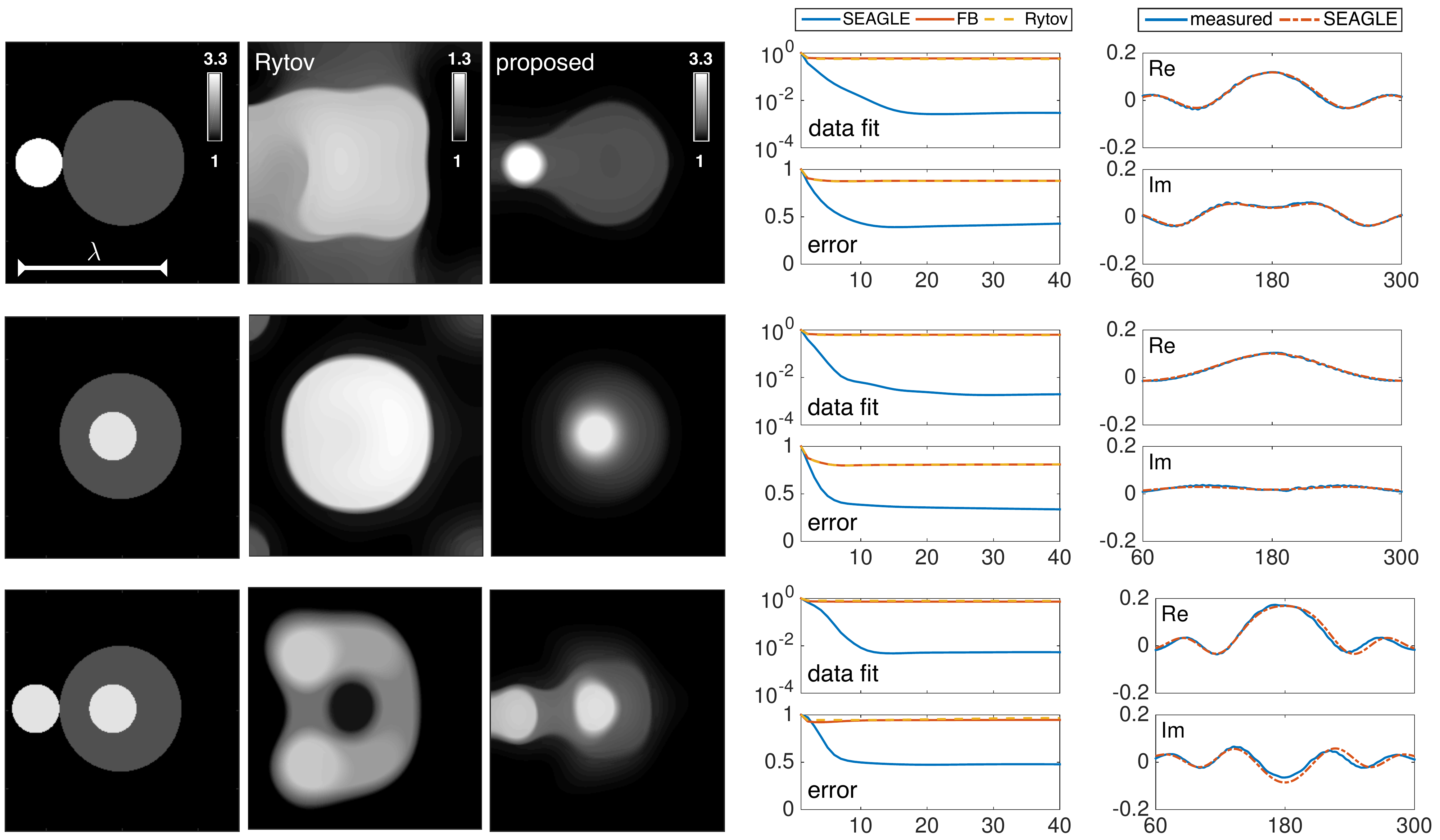}
\caption{\label{Fig:Experimental}
Reconstruction from an experimentally measured objects at 3 GHz, from top to bottom: \emph{FoamDielExtTM}, \emph{FoamDielIntTM}, and \emph{FoamTwinDielTM}. From left to right: ground truth; reconstruction using the Rytov approximation with TV regularization; reconstruction with the proposed method; evolution of normalized data-fit (top) and the normalized reconstruction error (bottom); and the true and predicted measurements for the transmission angle zero.}
\end{figure*}


We next validated the proposed technique for reconstructing the Shepp-Logan phantom in an ill-posed, strongly scattering, and compressive regime ($M = 25 \times 338$ and $N = 250 \times 250$). Specifically, we consider the setup in Fig.~\ref{Fig:Scheme} where the scattered wave measurements are generated with an FDTD simulator. The object is of size $84.9$~cm $\times$ $113$~cm and has a contrast of $20\%$. We place two linear detectors on either side of the phantom at a distance of $95.9$~cm from the center of the object. Each detector has $169$ sensors placed with a spacing of $3.84$~cm. The transmitters are positioned on a line $48.0$~cm  left to the left detector. They are spaced uniformly in azimuth with respect to the center of the phantom (every $5^\circ$ within $\pm60^\circ$). We set up a $120$~cm $\times$ $120$~cm square area for reconstructing the object, with pixel size $0.479$~cm. The wavelength of the illuminating light is $7.49$~cm.

Figure~\ref{fig:inverse} summarizes the performance of the proposed method, as well as that of the first Born and the Rytov approximations---the two baseline methods. All three approaches rely on FISTA with TV regularizer that was empirically set to $\tau = 1.5\times10^{-9}\|\ybf\|_{\ell_2}^2$ for the best performance. The order of our forward model is set to $K = 120$, but Algorithm~\ref{alg:forward} may terminate earlier when the objective function~\eqref{Eq:ForwardModel} is below $\delta_{\text{\tiny tol}} = 5\times10^{-7}\|\ubfin\|_{\ell_2}^2$. The figure additionally provides quantitative performance evaluation in terms of two quantities defined as
\begin{subequations}
\label{Eq:QuantitativeMetrics}
\begin{align}
&\text{normalized data fit} \defn \frac{\Dcal(\fbfhat)}{\Dcal(\bm{0})} = \frac{\|\zbf(\fbfhat)-\ybf\|_{\ell_2}^2}{\|\ybf\|_{\ell_2}^2}\\
&\text{normalized reconst. error} \defn \frac{\|\fbfhat-\fbf\|_2^{\ell_2}}{\|\fbf\|_2^{\ell_2}},
\end{align}
\end{subequations}
where $\fbf$ and $\fbfhat$ denote the true and estimated object, respectively.
Simulation results corroborate the benefit of using the proposed method for strongly scattering objects. It can be seen that, due to the ill-posed nature of the measurements, the reconstructed images suffer from missing frequency artifacts~\cite{Sheppard.Cogswell1990}. However, the proposed method is still able to accurately capture most features of the object while the linear methods fail to do so. Note also, that our method was initialized with the background value of the dielectric permittivity, $\epsilon_{\text{\tiny b}} = 1$, and that it takes fewer than $50$ FISTA iterations for converging to a stationary point (see convergence plot in Figure~\ref{fig:inverse}(f)).

\subsection{Validation on experimental data}

\begin{figure}[t]
\centering
\includegraphics[width=8.5cm]{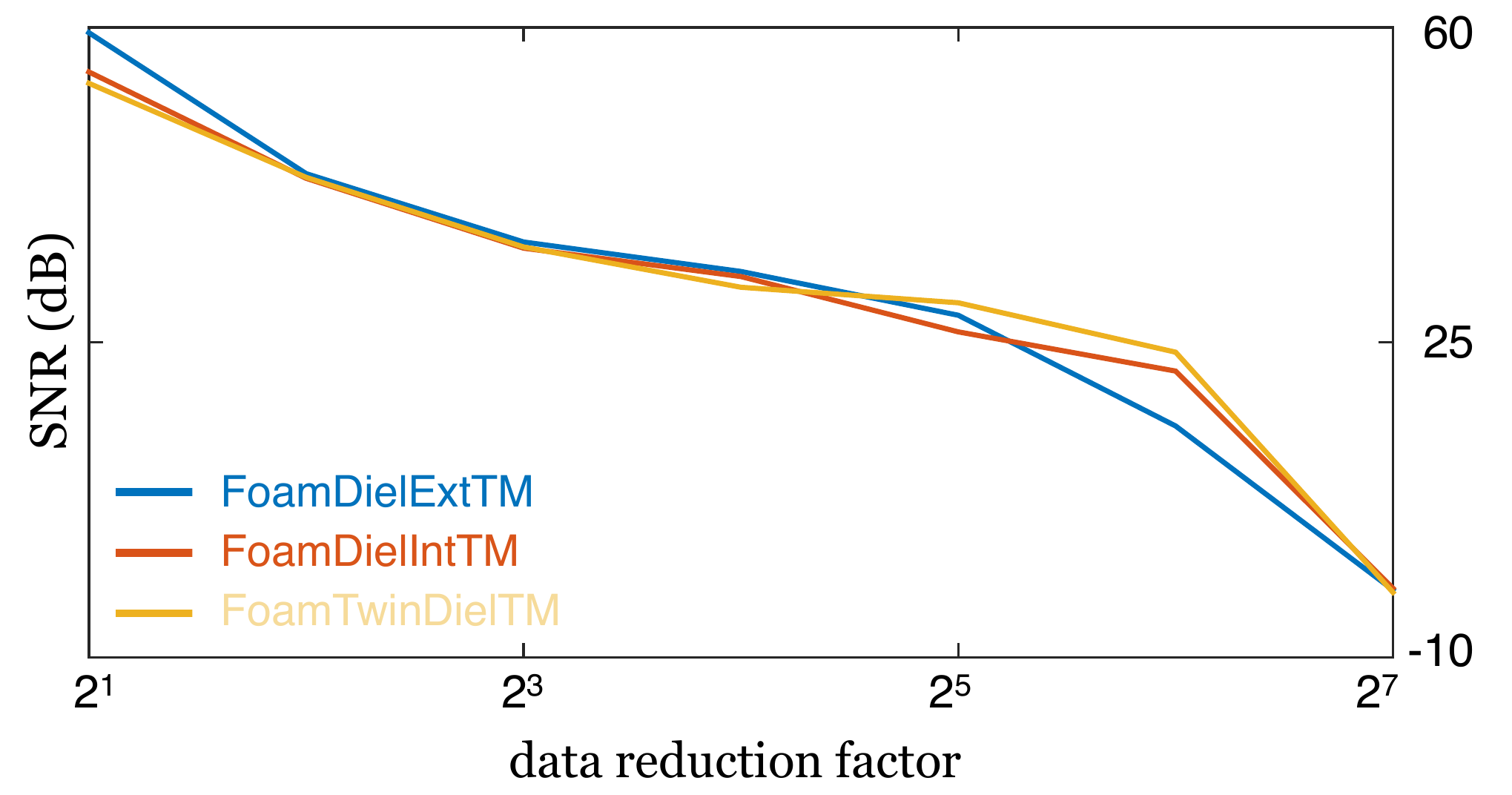}
\caption{\label{Fig:DataReduction}
Reconstruction quality (see text) of the proposed method at various values of the data-reduction factor.}
\end{figure}

We apply our method to three objects from the public dataset provided by the Fresnel institute~\cite{Geffrin.etal2005}: \textit{FoamDielExtTM}, \textit{FoamDielIntTM}, and \textit{FoamTwinDielTM}. These objects are placed in a region of size 15~cm $\times$ 15~cm  at the center of a circular rim of radius 1.67~m and measured using 360 detectors and 8 transmitters evenly distributed on the rim. The number of transmitters is increased to 18 for \textit{FoamTwinDielTM} and are also uniformly spaced. In all cases, only one transmitter is turned on at a time, while 241 detectors are used for each transmitter by excluding 119 detectors that are closest to the transmitter. While the full data contains multiple frequency measurements, we only use the data corresponding to the $3$ GHz. As before, we compare the result of our method with first Born and Rytov approximations that are also regularized with TV. We set the highest order of forward scattering to $K=200$ and the TV regularization to $\tau = 0.25\times10^{-8}\|\ybf\|_{\ell_2}^2$, and run the image formation algorithm for $40$ FISTA iterations. The reconstruction was initialized with the background value of the dielectric permittivity, which in this case corresponds to $\epsilon_{\text{\tiny b}} = 1$.

Figure~\ref{Fig:Experimental} summarizes the imaging results on the experimental data. The quantitative evaluation is performed using the same metrics defined in~\eqref{Eq:QuantitativeMetrics}. The results show that our method successfully captures the shape of the samples, as well as the value of the permittivity. Both first Born and Rytov approximations underestimate the permittivity. One can also see that the data-fit error for both of the linear forward models remain high as iterations progress. On the other hand, the object reconstructed by the proposed method closely agrees with the measured data (see rightmost column in Figure~\ref{Fig:Experimental}).

Figure~\ref{Fig:DataReduction} illustrates the performance of our method when using a limited number of measured data. In particular, we consider the reconstruction of the same three objects, but reduce the number of measurements for each transmission using regular downsampling by factors of $2, 4, 8, 16, 32, 64,$ and $128$. The full dataset consists of $8$ transmissions with $241$ measurements each; a factor of $128$ downsampling reduces to $8$ transmissions with $2$ measurements each. The size of the reconstructed image is set to $320 \times 320$ pixels. The reconstruction performance is quantified as
\begin{equation}
\text{SNR (dB) } \defn 10\log_{10} \left(\frac{\|\fbf_{\text{\tiny ref}}\|_{\ell_2}^2}{\|\fbfhat-\fbf_{\text{\tiny ref}}\|_{\ell_2}^2}\right),
\end{equation}
where $\fbf_{\text{\tiny ref}}$ is the reconstructed image with all the measured data (see Figure~\ref{Fig:Experimental}). The visual illustration is provided for \emph{FoamDielExtTM} in Figure~\ref{Fig:MainFigure}. This result highlights the stability of the proposed method to subsampling and experimental noise, even at highly nonlinear scattering scenarios.

Note that several other methods have been tested on this dataset~\cite{Abubakar.etal2005, Baussard2005, Crocco.etal2005, Dubois.etal2005, Feron.etal2005}. Qualitative comparison of our results in Figure~\ref{Fig:Experimental} with the results of those methods indicates that our approach achieves comparable performance using only a fraction of data (\emph{i.e.}, a single frequency with possible subsampling). Additionally, we observe a reliably stable and fast convergence starting from the initialization to the background permittivity, which is desirable in strongly scattering regimes. 


\section{Conclusion}
\label{Sec:Conclusion}

In conclusion, we have demonstrated a nonconvex optimization technique for solving nonlinear inverse scattering problems. We have applied the technique to simulated and experimentally measured data in microwave frequencies. The scattering was modeled as a series expansion with Nesterov's accelerated-gradient method. By structuring the expansion as a recursive feedforward network, we derived an error backpropagation formula for evaluating the gradient that can be used for iterative image reconstruction. The algorithm yields images of better quality than methods using linear forward models and is competitive with state-of-the-art inverse scattering approaches, tested on the same dataset. While the optimization problem is not convex, we have observed that the algorithm converges reliably within 100 iterations from a constant initialization of the permittivity. Our approach provides a promising framework for active correction of scattering in various applications and has the potential of significantly increasing the resolution and robustness when imaging strongly scattering objects.


\section{Total variation minimization}
\label{Sec:TVMinimization}

In this section, we review ideas and algorithms behind total variation (TV) regularized reconstruction. This section reviews the algorithm that was originally proposed by Beck and Teboulle~\cite{Beck.Teboulle2009a}.

Two common variants of TV regularizers are the \emph{anisotropic} TV regularizer
\begin{equation}
\label{Eq:AnisotropicTVApp}
\Rcal(\fbf) \defn \sum_{n = 1}^N \|[\Dbf\fbf]_n\|_{\ell_1} 
= \sum_{n = 1}^N \sum_{d = 1}^D |[\Dbf_d \fbf]_n|
\end{equation}
and the \emph{isotropic} TV regularizer
\begin{equation}
\label{Eq:IsotropicTVApp}
\Rcal(\fbf) 
\defn \sum_{n = 1}^N \|[\Dbf\fbf]_n\|_{\ell_2} = \sum_{n = 1}^N \sqrt{\sum_{d = 1}^D |[\Dbf_d\fbf]_n|^2}
\end{equation}
Here, $\Dbf: \R^N \rightarrow \R^{N \times D}$ is the discrete gradient operator and $D$ is the number of dimensions in the signal. The matrix $\Dbf_d$ denotes the finite difference operator along the dimension $d$
with appropriate boundary conditions (periodization, Neumann boundary conditions, etc.).

The anisotropic TV regularizer~\eqref{Eq:AnisotropicTVApp} can be interpreted as a sparsity-promoting $\ell_1$-penalty on the image gradient, while its isotropic counterpart~\eqref{Eq:IsotropicTVApp} as an $\ell_1$-penalty on the \emph{magnitudes} of the image gradient, which can also be viewed as a penalty promoting joint-sparsity of the gradient components. By promoting signals with sparse gradients, TV minimization recovers images that are piecewise-smooth, which means that they consist of smooth regions separated by sharp edges. Isotropic TV regularizer~\eqref{Eq:IsotropicTVApp} is rotation invariant, which makes it preferable in the context of image reconstruction. 

\begin{algorithm}[t]
\caption{FGP for evaluating $\fbf = \prox_{\tau \Rcal}(\zbf)$.}
\label{Algo:FGP}
\begin{algorithmic}[1]
\Statex{\textbf{input: } $\zbf \in \R^N$, $\tau > 0$.}
\Statex{\textbf{set: } $t \leftarrow 1$, $\gbftilde^0 \leftarrow \gbf^0$, $q_0  \leftarrow 1$, $\gamma \leftarrow 1/(12\tau)$}
	\Repeat
	\State $\gbf^t  \leftarrow  \proj_{\Gcal} \left(\gbftilde^{t-1} + \gamma \Dbf\left(\proj_\Fcal \left(\zbf-\tau\Dbf^\Trm\gbftilde^{t-1}\right)\right) \right)$
	\State $\fbf^t  \leftarrow \proj_\Fcal \left(\zbf - \tau \Dbf^\Trm\gbf^t\right) $
	\State $q_t  \leftarrow \frac{1}{2}\left(1 + \sqrt{1+4q_{t-1}^2}\right)$
	\State $\gbftilde^{t}  \leftarrow \gbf^t + ((q_{t-1}-1)/q_t)(\gbf^t - \gbf^{t-1})$
	\State $t \leftarrow t+1$
	\Until{stopping criterion}
	\Statex{\textbf{return:} $\fbf^t$}
\end{algorithmic}
\end{algorithm}

FISTA summarized in Algorithm~\ref{Algo:FISTA}, is a popular algorithm for solving inverse problems. FISTA relies on the efficient evaluation of the gradient $\nabla \Dcal$ and of the proximal operator
\begin{equation}
\label{Eq:Proximal}
\prox_{\tau \Rcal} (\zbf) \defn \argmin_{\fbf \in \Fcal} \left\{\frac{1}{2}\|\fbf - \zbf\|_{\ell_2}^2 + \tau \Rcal(\fbf)\right\}.
\end{equation}
The error backpropagation algorithm makes application of FISTA straightforward for solving inverse scattering with regularizers that admit closed form poximal operators such as the $\ell_1$-penalty. However, some regularizers including TV do not have closed form proximals and require an additional iterative algorithm for solving~\eqref{Eq:Proximal}.

Our implementation solves~\eqref{Eq:Proximal} with the dual approach that was proposed by Beck and Teboulle~\cite{Beck.Teboulle2009a}. The fast-gradient projection (FGP) approach, summarized in Algorithm~\ref{Algo:FGP}, is based on iteratively solving the dual optimization problem
\begin{equation}
\label{Eq:DualOptimization}
\gbfhat = \argmin_{\gbf \in \Gcal} \left\{\Qcal(\gbf)\right\},
\end{equation}
where
\begin{align}
\Qcal(\gbf) &\defn -\frac{1}{2}\|\zbf - \tau \Dbf^\Trm\gbf - \proj_\Fcal(\zbf-\tau\Dbf^\Trm\gbf)\|^2_{\ell_2}  \\
&\quad\quad+ \frac{1}{2}\|\zbf - \tau\Dbf^\Trm\gbf\|^2_{\ell_2}.\nonumber
\end{align}
Given the dual iterate $\gbf^t$, the corresponding primal iterate can be computed as
\begin{equation}
\fbf^t = \proj_\Fcal(\zbf - \tau \Dbf^\Trm\gbf^t).
\end{equation}
The operator $\proj_\Fcal$ represents an orthogonal projection onto the convex set $\Fcal$. For example, a projection onto $N$-dimensional cube
\begin{equation}
\Fcal \defn \left\{\fbf \in \R^N \,:\, a \leq f_n \leq b, \, \forall n \in [1, \dots, N] \right\},
\end{equation}
with bounds $a$, $b > 0$, is given by
\begin{equation}
[\proj_{\Fcal}(\fbf)]_n =
\begin{cases}
a       & \quad \text{if } f_n < a\\
f_n  & \quad \text{if } a \leq f_n \leq b \\
b & \quad \text{if } f_n > b,
\end{cases} 
\end{equation}
for all $n \in [1, \dots, N]$. 

The set $\Gcal \subseteq \R^{N \times D}$ in~\eqref{Eq:DualOptimization} depends on the variant of TV used for regularization. For anisotropic TV~\eqref{Eq:AnisotropicTVApp}, the set corresponds to 
\begin{equation}
\Gcal \defn \{ \gbf \in \R^{N \times D} : \|[\gbf]_n\|_{\ell_\infty} \leq 1, \forall n \in [1, \dots, N]\}
\end{equation}
with the corresponding projection
\begin{equation}
[\proj_\Gcal(\gbf)]_n = 
\begin{pmatrix} 
\frac{[\gbf_x]_n}{\max(1, |[\gbf_x]_n|)} \\[0.5em]
\frac{[\gbf_y]_n}{\max(1, |[\gbf_y]_n|)}
\end{pmatrix}, 
\end{equation}
for all $n \in [1, \dots, N]$. Similarly, for isotropic TV~\eqref{Eq:IsotropicTVApp}, the set corresponds to 
\begin{equation}
\Gcal \defn \{ \gbf \in \R^{N \times D} : \|[\gbf]_n\|_{\ell_2} \leq 1, \forall n \in [1, \dots, N]\}
\end{equation}
with the corresponding projection
\begin{equation}
[\proj_\Gcal(\gbf)]_n = \frac{[\gbf]_n}{\max\left(1, \|[\gbf]_n\|_{\ell_2}\right)}, 
\end{equation}
for all $n \in [1, \dots, N]$.

While the theoretical convergence of FISTA requires
the full convergence of inner Algorithm~\ref{Algo:FGP}, in practice, it is sufficient to run about $5$-$10$ iterations with an initializer that corresponds to the dual variable from the previous outer iteration. In our implementation, we thus fix the maximal number of inner iterations to $t_{\text{\tiny in}} = 10$
and enforce an additional stopping criterion based on measuring the relative change of the solution in two
successive iterations as $\|\gbf^t - \gbf^{t-1}\|_{\ell_2}/\|\gbf^{t-1}\|_{\ell_2} \leq \delta_{\text{in}}$, where $\delta_{\text{in}} = 10^{-4}$ in all the experiments. 


\section{Derivation of Error Backpropagation}
\label{Sec:EBP}

In this appendix, we provide the derivation of error backpropagation applied to our method. The method essentially computes the gradient of the data-fidelity term. This gradient is a key step of updating the scattering potential in solving the inverse problem. We now present the mathematical derivation of the gradient computation and relate it to Algorithm \ref{alg:backward}.

The inputs of the error-back propagation are the data mismatch and the intermediate variables ($\{\sbf^k\}$, $\{\gamma_k\}$, $\{\mu_k\}$, $\ubfhat = \ubf^K$) of the forward model computation, and the output is the gradient. Here we follow the differentiation conventions for vectors: $(\frac{\partial \ubf}{\partial \fbf})_{ij}=\frac{\partial \ubf_i}{\partial \fbf_j}$ and $(\grad_\fbf \ubf)_{ij}=[(\frac{\partial \ubf}{\partial \fbf})^\Hrm]_{ij}=\frac{\partial \ubf^*_j}{\partial \fbf_i}$. All boldface lower-case variables are column vectors.

Let us begin with the gradient of $\Dcal=\frac{1}{2}\|\zbf-\ybf\|^2_2$. 
\begin{align}
\grad_\fbf\Dcal =& \frac{1}{2}\grad_\fbf[(\zbf-\ybf)^\Hrm(\zbf-\ybf)] \notag\\
=& \frac{1}{2}[(\grad_\fbf\zbf)(\zbf-\ybf) + ((\zbf-\ybf)^\Hrm(\grad_\fbf \zbf)^\Hrm)^\Trm]\notag\\
=& \mathrm{Re}\left\{(\grad_\fbf\zbf)(\zbf-\ybf)\right\}. \label{eq:grad-D}
\end{align}
This can be evaluated by applying the chain rule to $\grad_\fbf\zbf$ and all the variables composing $\zbf$. The equations leading from the initialization all the way to $\zbf$ are listed below:
\begin{align*}
&\zbf = \ubf_\sub{in}+\mathbf{\Hbf}\diag\{\fbf\}\ubf^K\\
&\sbf^k = (1-\mu_k) \ubf^{k-1} + \mu_k\ubf^{k-2}\\
&\ubf^k = \sbf^k-\gamma_k\Abf^\Hrm(\Abf\sbf^k-\ubf_\sub{in}),
\end{align*}
for $k = 1,\ldots,K$, where $\Abf \defn \Ibf-\Gbf\diag\{\fbf\}$, and $\ubf^{-1} = \ubf^0$. It is worth noting that, while the step-size $\gamma_k$ also depends on $\fbf$, we ignore this dependency to simplify the computation. The rationale for this simplification is that the step-size can be replaced by a fixed one. Furthermore, in practice, $\gamma_k$ attains a stationary value for large enough $k$, which indicates that this simplification has a negligible effect on backpropagation. 

\subsection{Initialization of backpropagation}
The initialization in Algorithm~\ref{alg:backward} is obtained by differentiating the first of the above equations with respect to $\fbf$. With ${\diag\{\fbf\}\ubf=\diag\{\ubf\}\fbf}$, we have
\begin{align}
\grad_\fbf \zbf &= \left[\mathbf{\Hbf} \frac{\partial \fbf}{\partial \fbf} \diag\{\ubf^K\} + \mathbf{\Hbf}\diag\{\fbf\} \frac{\partial \ubf^K}{\partial\fbf} \right]^\Hrm \notag\\
&= \diag\{\ubf^K\}^\Hrm \mathbf{\Hbf}^\Hrm + (\grad_\fbf \ubf^K)\diag\{\fbf\}^\Hrm\mathbf{\Hbf}^\Hrm \label{eq:grad-uhat}
\end{align}
The first term gives the remainder that contributes to the final result while the second term gives the vector that multiplies with $\grad_\fbf \ubf^K$. For convenience, we define two sets of vectors: \begin{itemize}
\item $\qbf^k$: the vector that multiplies with $\grad_\fbf \ubf^k$
\item $\rbf^k$: the remainder before computing $(\grad_\fbf \ubf^k)\qbf^k$.
\end{itemize}
In addition, due to the acceleration step in the forward computation, we expect subsequent $\qbf^{k-1}$ to have a contribution from $(\grad_\fbf \ubf^{k+1})\qbf^{k+1}$ in addition to the contribution from its direct neighbor $(\grad_\fbf \ubf^{k})\qbf^{k}$. This leads to the third set of vectors: \begin{itemize}
\item $\pbf^k$: the explicit contribution of $(\grad_\fbf \ubf^{k+1})\qbf^{k+1}$ to $\qbf^{k-1}$.
\end{itemize}
Finally, multiplying (\ref{eq:grad-uhat}) with $(\zbf-\ybf)$ we identify
\begin{align}
\rbf^K &= \diag\{\ubf^K\}^\Hrm \mathbf{\Hbf}^\Hrm (\zbf-\ybf) \label{eq:rK}\\
\qbf^K &= \diag\{\fbf\}^\Hrm\mathbf{\Hbf}^\Hrm (\zbf-\ybf). \label{eq:qK}
\end{align}
Since there is no term multiplying with $\grad_\fbf \ubf^{K-1}$ explicitly (hence nothing to pass to $\qbf^{K-1}$), we have
\begin{equation}\label{eq:pK}
\pbf^K = \mathbf{0}.
\end{equation}

\subsection{Recursive updates for $\ubf^k$}
The computation of $(\grad_\fbf \ubf^k)\qbf^k$ is the key step in error-back propagation. We evaluate the gradient $\grad_\fbf \ubf^k$ by taking the Hermitian of the derivative, and the multiplication with $\qbf^k$ follows. The result should be passed onto another gradient with smaller $k$. Before we start, let us write out the gradient of $\sbf^k$ which is straightforward from its definition, 
\begin{equation} \label{eq:grad-yk}
\grad_\fbf \sbf^k = (1-\mu_k) \grad_\fbf\ubf^{k-1} + \mu_k \grad_\fbf \ubf^{k-2}.
\end{equation}
The derivative of $\ubf^k$ is
\begin{equation}\label{eq:diff-uk} \begin{split}
\frac{\partial\ubf^k}{\partial\fbf} &= 
    -\gamma_k\frac{\partial\Abf^\Hrm}{\partial\fbf}(\Abf\sbf^k-\ubf_\sub{in})
    -\gamma_k\Abf^\Hrm \frac{\partial\Abf}{\partial\fbf}\sbf^k\\
    &\quad +(\Ibf-\gamma_k\Abf^\Hrm\Abf)\frac{\partial\sbf^k}{\partial\fbf}.
\end{split}\end{equation}
The first term becomes
\begin{align*}
-\gamma_k\frac{\partial\Abf^\Hrm}{\partial\fbf}(\Abf\sbf^k-\ubf_\sub{in})
&= \gamma_k\left(\frac{\partial}{\partial\fbf}\diag\{\fbf\}\right)\mathbf{G}^\Hrm(\Abf\sbf^k-\ubf_\sub{in}) \notag\\
&= \gamma_k\diag\{\mathbf{G}^\Hrm(\Abf\sbf^k-\ubf_\sub{in})\},
\end{align*}
and the second term becomes
\begin{align*}
-\gamma_k\Abf^\Hrm \frac{\partial\Abf}{\partial\fbf}\sbf^k
&= \gamma_k \Abf^\Hrm \mathbf{G} \left(\frac{\partial}{\partial\fbf}\diag\{\fbf\}\right) \sbf^k \notag\\
&= \gamma_k \Abf^\Hrm \mathbf{G} \diag\{\sbf^k\}.
\end{align*}
By taking Hermitian transpose of \eqref{eq:diff-uk}, we have
\begin{align} 
\grad_\fbf \ubf^k = \gamma_k \mathbf{T}^k + (\grad_\fbf \sbf^k) \mathbf{S}^k \label{eq:grad-uk} 
\end{align}
where
\begin{align} 
\mathbf{T}^k &= \diag\{\mathbf{G}^\Hrm(\Abf\sbf^k-\ubf_\sub{in})\}^\Hrm 
    + \diag\{\sbf^k\}^\Hrm \mathbf{G}^\Hrm \Abf \\
\mathbf{S}^k &= (\Ibf-\gamma_k\Abf^\Hrm\Abf)^\Hrm = \Ibf-\gamma_k\Abf^\Hrm\Abf.
\end{align}
By multiplying \eqref{eq:grad-uk} with $\qbf^k$ and substituting with \eqref{eq:grad-yk}, we obtain the expression for $(\grad_\fbf \ubf^k)\qbf^k$,
\begin{equation}\label{eq:grad-uk-qk}
\begin{split}
(\grad_\fbf \ubf^k)\qbf^k &= \gamma_k \mathbf{T}^k \qbf^k  \\
&\quad + (\grad_\fbf\ubf^{k-1}) \left[(1-\mu_k)\mathbf{S}^k\qbf^k\right]  \\
&\quad + (\grad_\fbf \ubf^{k-2}) \left[\mu_k\mathbf{S}^k\qbf^k\right].
\end{split}\end{equation}
Note that because we set $\ubf^{-1} = \ubf^0$,
\begin{equation}\label{eq:grad-u1-q1}
(\grad_\fbf \ubf^1)\qbf^1 = \gamma_1 \mathbf{T}^1 \qbf^1 + (\grad_\fbf\ubf^{0}) \mathbf{S}^1\qbf^1.
\end{equation}

\subsection{Error backpropagation equations}
From equations (\ref{eq:grad-uhat}) to (\ref{eq:pK}), we have
\begin{align}
(\grad_\fbf\zbf)(\zbf-\ybf) &= \rbf^K + (\grad_\fbf \ubf^K) \qbf^K + (\grad_\fbf \ubf^{K-1}) \pbf^K \label{eq:ebpK}
\end{align}
Substituting \eqref{eq:grad-uk-qk} into \eqref{eq:ebpK}, we have the following expressions
\begin{align}
(\grad_\fbf\zbf)(\zbf-\ybf)
=&\,\rbf^K + (\grad_\fbf \ubf^K) \qbf^K + (\grad_\fbf \ubf^{K-1}) \pbf^K \notag \\
=&\,\rbf^{K-1} + (\grad_\fbf \ubf^{K-1}) \qbf^{K-1} + (\grad_\fbf \ubf^{K-2}) \pbf^{K-1} \notag \\
=&\, \ldots\notag \\
=&\,\rbf^1 + (\grad_\fbf \ubf^1) \qbf^1 + (\grad_\fbf \ubf^{0}) \pbf^1 \label{eq:ebp1}
\end{align}
and the recursion relations for $k=2 \ldots K$
\begin{align}
\rbf^{k-1} &= \rbf^k + \gamma_k \mathbf{T}^k \qbf^k \label{eq:r-recur}\\
\qbf^{k-1} &= \pbf^k + (1-\mu_k)\mathbf{S}^k\qbf^k \label{eq:q-recur}\\
\pbf_{k-1} &= \mu_k\mathbf{S}^k\qbf^k. \label{eq:p-recur}
\end{align}
For the case $k=1$, or namely $\rbf_0$ and $\qbf_0$ (note that $\pbf_0$ does not exist due to $\ubf^{-1} = \ubf^0$), we plug \eqref{eq:grad-u1-q1} into \eqref{eq:ebp1},
\begin{align*}
(\grad_\fbf\zbf)(\zbf-\ybf) &= \rbf^1 + \gamma_1 \mathbf{T}^1 \qbf^1 + (\grad_\fbf\ubf^{0})\left[\mathbf{S}^1\qbf^1 + \pbf^1\right].
\end{align*}
Hence we have 
\begin{align}
\rbf^0 &= \rbf^1 + \gamma_1 \mathbf{T}^1 \qbf^1 \label{eq:r0}\\
\qbf^0 &= \pbf^1 + \mathbf{S}^1\qbf^1. \label{eq:q0}
\end{align}

In the initialization of our forward model, $\ubf^0$ is the incident field and does not depend on $\fbf$. Therefore $\grad_\fbf \ubf^0 = \mathbf{0}$ and the gradient of data-fidelity is 
\begin{equation}
\grad_\fbf \Dcal = \mathrm{Re}\left\{(\grad_\fbf\zbf)(\zbf-\ybf)\right\} = \mathrm{Re}\left\{ \rbf^0 \right\}.
\end{equation}
We summarize these recursion relations of error backpropagation in Algorithm \ref{alg:backward} of the main text.


\section{Analytic Solutions to Scattering}
\label{Sec:Analytic}

In this section, our aim is to present the analytic expressions for scalar electric fields resulting from a point source outside a dielectric sphere in 2D and 3D (strictly speaking, the 2D case should be understood as an infinitely long line source illuminating a cylinder parallel to it and looking at the cross-section). A sketch of the derivation is provided after the actual expressions. A more complete description can be found in a number of standard textbooks such as~\cite{Jackson1999}.

\subsection{Expressions}
Consider a sphere of a radius $r_\sub{sph}$ and a refractive index $n = \sqrt{\epsilon}$. The source is located $r_\sub{s}$ distance away from the center of the sphere and the wavenumber of the source outside the sphere is $k_b$.

\medskip\noindent\emph{2D case:}\\
We consider the polar coordinates 
\[\rbf=(r\cos\theta,r\sin\theta),\]
and, without loss of generality, assume that the source is at $\theta_\sub{s}=0$. The field can be expressed as
\begin{equation}\label{eq:analytical:2Dsol1}
    E(\rbf;r_\sub{s}) = \sum_{m=-\infty}^{\infty}R_m(r,r_\sub{s})\frac{e^{\jrm m\theta}}{2\pi}
\end{equation}
where $\rho = k_b r$, $\rho_\sub{sph} = k_b r_\sub{sph}$ and $\rho_\sub{s} = k_b r_\sub{s}$,
\begin{align}
&R_m(r,r_\sub{s}) \notag\\
&= \begin{cases}
a_m J_m(n\rho) H_m^{(1)}(\rho_\sub{s}),   & \mbox{ $r<r_\sub{sph}$}\\
(b_m J_m(\rho) + c_m Y_m(\rho)) H_m^{(1)}(\rho_\sub{s}),   & \mbox{ $r_\sub{sph}\leq r<r_\sub{s}$}\\
(b_m J_m(\rho_\sub{s}) + c_m Y_m(\rho_\sub{s})) H_m^{(1)}(\rho),     & \mbox{ $r_\sub{s}\leq r$}
\end{cases}\\
&a_m = \frac{-1}{\rho_\sub{sph}\Delta_m}\\
&b_m = \frac{-\pi}{2\Delta_m} \left|\begin{array}{cc}  
        J_{m}(n\rho_\sub{sph})  &  nJ_{m-1}(n\rho_\sub{sph}) \\
        Y_{m}(\rho_\sub{sph})   &  Y_{m-1}(\rho_\sub{sph})
    \end{array}\right|\\
&c_m =  \frac{\pi}{2\Delta_m} \left|\begin{array}{cc}  
        J_{m}(n\rho_\sub{sph})  &  nJ_{m-1}(n\rho_\sub{sph}) \\
        J_{m}(\rho_\sub{sph})   &  J_{m-1}(\rho_\sub{sph})
    \end{array}\right|\\
&\Delta_m = \left|\begin{array}{cc}  
        J_{m}(n\rho_\sub{sph})      &  nJ_{m-1}(n\rho_\sub{sph}) \\
        H^{(1)}_{m}(\rho_\sub{sph}) &  H^{(1)}_{m-1}(\rho_\sub{sph})
    \end{array}\right| \label{eq:analytical:2Dsol2}
\end{align}
and $J_m$ and $Y_m$ are the $m$'th order Bessel functions of the first kind and the second kind, and $H^{(1)}_m=J_m+\jrm Y_m$ is the $m$'th order Hankel's function of the first kind.

\medskip\noindent\emph{3D case:}\\
We consider the spherical coordinates
\[\rbf=(r\sin\theta\cos\phi,r\sin\theta\sin\phi,r\cos\theta),\] 
and, without loss of generality, assume that the source has zenith angle $\theta_\sub{s}=0$ and azimuthal angle $\phi_\sub{s}=0$. The field then reads
\begin{equation}\label{eq:analytical:3Dsol1}
E(\rbf;r_\sub{s}) = \sum_{l=0}^{\infty}R_l(r,r_\sub{s}) \left(\frac{2l+1}{4\pi}\right)P_l(\cos\theta)
\end{equation}
where, with $\rho = k_b r$, $\rho_\sub{sph} = k_b r_\sub{sph}$ and $\rho_\sub{s} = k_b r_\sub{s}$,
\begin{align}
R_l(r,r_\sub{s}) &= \begin{cases}
A_l  j_l(n\rho)  h_l^{(1)}(\rho_\sub{s}),   & \mbox{ $r<r_\sub{sph}$}\\
(B_l j_l(\rho) + C_l n_l(\rho)) h_l^{(1)}(\rho_\sub{s}),   & \mbox{ $r_\sub{sph}\leq r<r_\sub{s}$}\\
(B_l j_l(\rho_\sub{s}) + C_l n_l(\rho_\sub{s})) h_l^{(1)}(\rho),     & \mbox{ $r_\sub{s}\leq r$}
\end{cases}\\
A_m &= \frac{k_b}{\rho^2_\sub{sph} D_m}\\
B_m &= \frac{-k_b}{D_m} \left|\begin{array}{cc}  
        j_{l}(n\rho_\sub{sph})  &  nj_{l+1}(n\rho_\sub{sph}) \\
        n_{l}(\rho_\sub{sph})   &  n_{l+1}(\rho_\sub{sph})
    \end{array}\right|\\
C_m &=  \frac{k_b}{D_m} \left|\begin{array}{cc}  
        j_{l}(n\rho_\sub{sph})  &  nj_{l+1}(n\rho_\sub{sph}) \\
        j_{l}(\rho_\sub{sph})   &  j_{l+1}(\rho_\sub{sph})
    \end{array}\right|\\
D_m &= \left|\begin{array}{cc}  
        j_{l}(n\rho_\sub{sph})      &  nj_{l+1}(n\rho_\sub{sph}) \\
        h^{(1)}_{l}(\rho_\sub{sph}) &  h^{(1)}_{l+1}(\rho_\sub{sph})
    \end{array}\right|
\end{align}
and $j_l$ and $n_l$ are the $l$'th order spherical Bessel function of the first kind and the second kind, $h^{(1)}_l=j_l+\jrm n_l$ is the $l$'th order of spherical Hankel function of the first kind, and $P_l(x)$ is the Legendre polynomial defined as
\begin{equation}\label{eq:analytical:3Dsol2}
P_l(x) = \frac{1}{2^l l!}\frac{d^l}{dx^l}(x^2+1)^l.
\end{equation}

\subsection{Helmholtz equation}

The Helmholtz equation for a point source is \begin{equation}\label{eq:analytical:helmholtz}
\grad_\rbf^2 E(\rbf,\rbf_\sub{s})+k^2(\rbf) E(\rbf,\rbf_\sub{s}) = -\delta(\rbf-\rbf_\sub{s})
\end{equation}
where $E$ is the complex electric field at position $\rbf$ when the point source is at $\rbf_\sub{s}$ and $k^2(\rbf)$ is defined as
\begin{equation}
k^2(\rbf) = k^2(\|\rbf\|_2) = \begin{cases}
 n^2k_b^2, & \mathrm{for}~\|\rbf\|_2 < r_\sub{sph} \\  
 k_b^2, & \mathrm{for}~\|\rbf\|_2 > r_\sub{sph}   
 \end{cases}.
\end{equation}
Note that both 2D and 3D cases follow the same form. Their difference is that $\rbf$ and $\rbf_\sub{s}$ have 2 or 3 coordinates.

\subsection{Derivation for 2D case}

We consider the polar coordinates, assume $\theta_\sub{s}=0$ without loss of generality, and use an ansatz for the electric field in~\eqref{eq:analytical:2Dsol1}. With the Laplacian in the polar coordinate and the following expansion of a 2D delta function~\cite{Jackson1999}
\begin{equation}\label{eq:analytical:2Dderive2}
\delta(\rbf-\rbf_\sub{s})=\frac{1}{r}\delta(r-r_\sub{s})\frac{1}{2\pi}\sum_{m=-\infty}^{\infty}e^{jm(\theta-\theta_\sub{s})},
\end{equation}
eq.~\eqref{eq:analytical:helmholtz} becomes a sequence of equations on $R_m(r,r_\sub{s})$
\begin{equation}
\begin{split}\label{eq:analytical:2Dderive3}
&\frac{\partial}{\partial r}\left(r\frac{\partial}{\partial r}R_m(r,r_\sub{s})\right) + \left(rk^2(r)-\frac{m^2}{r}\right)R_m(r,r_\sub{s}) \\
=& -\delta(r-r_\sub{s})
\end{split}
\end{equation}
for each $m$. Each equations is a Bessel differential equation so the solution can be composed of Bessel functions of order $m$. The boundary conditions for $R_m$ are as follows
\begin{enumerate}
\item finite value at $r=0$
\item only outgoing component at $r=\infty$
\item continuous and first-derivative-continuous at $r=r_\sub{sph}$
\item continuous at $r=r_\sub{s}$
\item $\left.\frac{\partial R_m}{\partial r}\right|_{r_\sub{s}^+}-\left.\frac{\partial R_m}{\partial r}\right|_{r_\sub{s}^-}=-\frac{1}{r_\sub{s}}$ (integrate (\ref{eq:analytical:2Dderive3}) around $r_\sub{s}$)
\end{enumerate}
The above condition and equations lead to Eqs.~(\ref{eq:analytical:2Dsol1})-(\ref{eq:analytical:2Dsol2}).

\subsection{Derivation for 3D case}
We consider the spherical coordinates and assume that the source lies on the zenith axis. The ansatz for the electric field is~\eqref{eq:analytical:3Dsol1}, the expansion of a 3D delta function is
\begin{align}
\delta(\rbf-\rbf_\sub{s})&=\frac{1}{r^2}\delta(r-r_\sub{s}) \sum_{l=0}^\infty \sum_{m=-l}^l Y_l^m(\theta,\phi) Y_l^m(0,0) \notag\\
&=\frac{1}{r^2}\delta(r-r_\sub{s}) \sum_{l=0}^\infty \left(\frac{2l+1}{4\pi}\right)P_l(\cos\theta)
\label{eq:analytical:3Dderive2}
\end{align}
and eq.~\eqref{eq:analytical:helmholtz} becomes
\begin{equation}\begin{split}\label{eq:analytical:3Dderive3}
&\frac{\partial}{\partial r}\left(r^2\frac{\partial}{\partial r}R_l(r,r_\sub{s})\right) + \left(k^2(r)r^2-l(l+1)\right)R_l(r,r_\sub{s}) \\
=& -\delta(r-r_\sub{s})
\end{split}
\end{equation}
for each $l$. These equations are spherical Bessel equations and there are corresponding spherical Bessel functions to compose the solution. The boundary conditions for the solution are the same as listed above except the last one becoming
$$\left.\frac{\partial R_l}{\partial r}\right|_{r_\sub{s}^+}-\left.\frac{\partial R_l}{\partial r}\right|_{r_\sub{s}^-}=-\frac{1}{r^2_\sub{s}}. \mbox{ (integrate~\eqref{eq:analytical:3Dderive3} around $r_\sub{s}$)}$$


\bibliographystyle{IEEEtran}

\begin{thebibliography}{10}
\providecommand{\url}[1]{#1}
\csname url@samestyle\endcsname
\providecommand{\newblock}{\relax}
\providecommand{\bibinfo}[2]{#2}
\providecommand{\BIBentrySTDinterwordspacing}{\spaceskip=0pt\relax}
\providecommand{\BIBentryALTinterwordstretchfactor}{4}
\providecommand{\BIBentryALTinterwordspacing}{\spaceskip=\fontdimen2\font plus
\BIBentryALTinterwordstretchfactor\fontdimen3\font minus
  \fontdimen4\font\relax}
\providecommand{\BIBforeignlanguage}[2]{{%
\expandafter\ifx\csname l@#1\endcsname\relax
\typeout{** WARNING: IEEEtran.bst: No hyphenation pattern has been}%
\typeout{** loaded for the language `#1'. Using the pattern for}%
\typeout{** the default language instead.}%
\else
\language=\csname l@#1\endcsname
\fi
#2}}
\providecommand{\BIBdecl}{\relax}
\BIBdecl

\bibitem{Kak.Slaney1988}
A.~C. Kak and M.~Slaney, \emph{Principles of Computerized Tomographic
  Imaging}.\hskip 1em plus 0.5em minus 0.4em\relax {IEEE}, 1988.

\bibitem{Wolf1969}
E.~Wolf, ``Three-dimensional structure determination of semi-transparent
  objects from holographic data,'' \emph{Opt. Commun.}, vol.~1, no.~4, pp.
  153--156, September/October 1969.

\bibitem{Devaney1981}
A.~J. Devaney, ``Inverse-scattering theory within the {R}ytov approximation,''
  \emph{Opt. Lett.}, vol.~6, no.~8, pp. 374--376, August 1981.

\bibitem{Born.Wolf1999}
M.~Born and E.~Wolf, \emph{Principles of Optics}, 7th~ed.\hskip 1em plus 0.5em
  minus 0.4em\relax Cambridge Univ. Press, 1999.

\bibitem{Candes.etal2006}
E.~J. Cand{\`e}s, J.~Romberg, and T.~Tao, ``Robust uncertainty principles:
  Exact signal reconstruction from highly incomplete frequency information,''
  \emph{IEEE Trans. Inf. Theory}, vol.~52, no.~2, pp. 489--509, February 2006.

\bibitem{Donoho2006}
D.~L. Donoho, ``Compressed sensing,'' \emph{IEEE Trans. Inf. Theory}, vol.~52,
  no.~4, pp. 1289--1306, April 2006.

\bibitem{Rudin.etal1992}
L.~I. Rudin, S.~Osher, and E.~Fatemi, ``Nonlinear total variation based noise
  removal algorithms,'' \emph{Physica D}, vol.~60, no. 1--4, pp. 259--268,
  November 1992.

\bibitem{Bronstein.etal2002}
M.~M. Bronstein, A.~M. Bronstein, M.~Zibulevsky, and H.~Azhari,
  ``Reconstruction in diffraction ultrasound tomography using nonuniform
  {FFT},'' \emph{IEEE Trans. Med. Imag.}, vol.~21, no.~11, pp. 1395--1401,
  November 2002.

\bibitem{Lim.etal2015}
J.~W. Lim, K.~R. Lee, K.~H. Jin, S.~Shin, S.~E. Lee, Y.~K. Park, and J.~C. Ye,
  ``Comparative study of iterative reconstruction algorithms for missing cone
  problems in optical diffraction tomography,'' \emph{Opt. Express}, vol.~23,
  no.~13, pp. 16\,933--16\,948, June 2015.

\bibitem{Sung.Dasari2011}
Y.~Sung and R.~R. Dasari, ``Deterministic regularization of three-dimensional
  optical diffraction tomography,'' \emph{J. Opt. Soc. Am. A}, vol.~28, no.~8,
  pp. 1554--1561, August 2011.

\bibitem{Boyd.Vandenberghe2004}
S.~Boyd and L.~Vandenberghe, \emph{Convex Optimization}.\hskip 1em plus 0.5em
  minus 0.4em\relax Cambridge Univ. Press, 2004.

\bibitem{Nocedal.Wright2006}
J.~Nocedal and S.~J. Wright, \emph{Numerical Optimization}, 2nd~ed.\hskip 1em
  plus 0.5em minus 0.4em\relax Springer, 2006.

\bibitem{Bioucas-Dias.Figueiredo2007}
J.~M. Bioucas-Dias and M.~A.~T. Figueiredo, ``A new {T}w{IST}: {T}wo-step
  iterative shrinkage/thresholding algorithms for image restoration,''
  \emph{IEEE Trans. Image Process.}, vol.~16, no.~12, pp. 2992--3004, December
  2007.

\bibitem{Beck.Teboulle2009a}
A.~Beck and M.~Teboulle, ``Fast gradient-based algorithm for constrained total
  variation image denoising and deblurring problems,'' \emph{IEEE Trans. Image
  Process.}, vol.~18, no.~11, pp. 2419--2434, November 2009.

\bibitem{Chen.Stamnes1998}
B.~Chen and J.~J. Stamnes, ``Validity of diffraction tomography based on the
  first born and the first rytov approximations,'' \emph{Appl. Opt.}, vol.~37,
  no.~14, pp. 2996--3006, May 1998.

\bibitem{Ntziachristos2010}
V.~Ntziachristos, ``Going deeper than microscopy: the optical imaging frontier
  in biology,'' \emph{Nat. Methods}, vol.~7, no.~8, pp. 603--614, August 2010.

\bibitem{Geffrin.etal2005}
J.-M. Geffrin, P.~Sabouroux, and C.~Eyraud, ``Free space experimental
  scattering database continuation: experimental set-up and measurement
  precision,'' \emph{Inv. Probl.}, vol.~21, no.~6, pp. S117--?S130, 2005.

\bibitem{Nesterov1983}
Y.~E. Nesterov, ``A method for solving the convex programming problem with
  convergence rate ${O}(1/k^2)$,'' \emph{Dokl. Akad. Nauk {SSSR}}, vol. 269,
  pp. 543--547, 1983, (in Russian).

\bibitem{Sharpe.etal2002}
J.~Sharpe, U.~Ahlgren, P.~Perry, B.~Hill, A.~Ross, J.~Hecksher-S{\o}rensen,
  R.~Baldock, and D.~Davidson, ``Optical projection tomography as a tool for
  {3D} microscopy and gene expression studies,'' \emph{Science}, vol. 296, no.
  5567, pp. 541--545, April 2002.

\bibitem{Choi.etal2007}
W.~Choi, C.~Fang-Yen, K.~Badizadegan, S.~Oh, N.~Lue, R.~R. Dasari, and M.~S.
  Feld, ``Tomographic phase microscopy,'' \emph{Nat. Methods}, vol.~4, no.~9,
  pp. 717--719, September 2007.

\bibitem{Lauer2002}
V.~Lauer, ``New approach to optical diffraction tomography yielding a vector
  equation of diffraction tomography and a novel tomographic microscope,''
  \emph{J. Microsc.}, vol. 205, no.~2, pp. 165--176, 2002.

\bibitem{Sung.etal2009}
Y.~Sung, W.~Choi, C.~Fang-Yen, K.~Badizadegan, R.~R. Dasari, and M.~S. Feld,
  ``Optical diffraction tomography for high resolution live cell imaging,''
  \emph{Opt. Express}, vol.~17, no.~1, pp. 266--277, December 2009.

\bibitem{Kim.etal2014}
T.~Kim, R.~Zhou, M.~Mir, S.~Babacan, P.~Carney, L.~Goddard, and G.~Popescu,
  ``White-light diffraction tomography of unlabelled live cells,'' \emph{Nat.
  Photonics}, vol.~8, pp. 256--263, March 2014.

\bibitem{Ralston.etal2006}
T.~S. Ralston, D.~L. Marks, P.~S. Carney, and S.~A. Boppart, ``Inverse
  scattering for optical coherence tomography,'' \emph{J. Opt. Soc. Am. A},
  vol.~23, no.~5, pp. 1027--1037, May 2006.

\bibitem{Davis.etal2007}
B.~J. Davis, S.~C. Schlachter, D.~L. Marks, T.~S. Ralston, S.~A. Boppart, and
  P.~S. Carney, ``Nonparaxial vector-field modeling of optical coherence
  tomography and interferometric synthetic aperture microscopy,'' \emph{J. Opt.
  Soc. Am. A}, vol.~24, no.~9, pp. 2527--2542, September 2007.

\bibitem{Brady.etal2009}
D.~J. Brady, K.~Choi, D.~L. Marks, R.~Horisaki, and S.~Lim, ``Compressive
  holography,'' \emph{Opt. Express}, vol.~17, no.~15, pp. 13\,040--13\,049,
  2009.

\bibitem{Tian.etal2010}
L.~Tian, N.~Loomis, J.~A. Dominguez-Caballero, and G.~Barbastathis,
  ``Quantitative measurement of size and three-dimensional position of
  fast-moving bubbles in air?water mixture flows using digital holography,''
  \emph{Appl. Opt.}, vol.~49, no.~9, pp. 1549--1554, March 2010.

\bibitem{Chen.etal2015a}
W.~Chen, L.~Tian, S.~Rehman, Z.~Zhang, H.~P. Lee, and G.~Barbastathis,
  ``Empirical concentration bounds for compressive holographic bubble imaging
  based on a {M}ie scattering model,'' \emph{Opt. E}, vol.~23, no.~4, p.
  February, 2015.

\bibitem{Jol2009}
H.~M. Jol, Ed., \emph{Ground Penetrating Radar: Theory and Applications}.\hskip
  1em plus 0.5em minus 0.4em\relax Amsterdam: Elsevier, 2009.

\bibitem{Leigsnering.etal2014a}
M.~Leigsnering, F.~Ahmad, M.~Amin, and A.~Zoubir, ``Multipath exploitation in
  through-the-wall radar imaging using sparse reconstruction,'' \emph{IEEE
  Trans. Aerosp. Electron. Syst.}, vol.~50, no.~2, pp. 920--939, April 2014.

\bibitem{Liu.etal2016a}
D.~Liu, U.~S. Kamilov, and P.~T. Boufounos, ``Compressive tomographic radar
  imaging with total variation regularization,'' in \emph{Proc. {IEEE} 4th
  International Workshop on Compressed Sensing Theory and its Applications to
  Radar, Sonar, and Remote Sensing ({CoSeRa} 2016)}, Aachen, Germany, September
  19-22, 2016, pp. 120--123.

\bibitem{Tijhuis1989}
A.~G. Tijhuis, ``{B}orn-type reconstruction of material parameters of an
  inhomogeneous, lossy dielectric slab from reflected-field data,'' \emph{Wave
  Motion}, vol.~11, no.~2, pp. 151--173, May 1989.

\bibitem{Wang.Chew1989}
Y.~M. Wang and W.~Chew, ``An iterative solution of the two-dimensional
  electromagnetic inverse scattering problem,'' \emph{Int. J. Imag. Syst
  Tech.}, vol.~1, pp. 100--108, 1989.

\bibitem{Chew.Wang1990}
W.~C. Chew and Y.~M. Wang, ``Reconstruction of two-dimensional permittivity
  distribution using the distorted {B}orn iterative method,'' \emph{IEEE Trans.
  Med. Imag.}, vol.~9, no.~2, pp. 218--225, June 1990.

\bibitem{Kleinman.etal1990}
R.~E. Kleinman, G.~F. Roach, and P.~M. {van den Berg}, ``Convergent {B}orn
  series for large refractive indices,'' \emph{J. Opt. Soc. Am. A}, vol.~7,
  no.~5, pp. 890--897, May 1990.

\bibitem{Haeberle.etal2010}
O.~Haeberl\'e, K.~Belkebir, H.~Giovaninni, and A.~Sentenac, ``Tomographic
  diffractive microscopy: basic, techniques, and perspectives,'' \emph{J. Mod.
  Opt.}, vol.~57, no.~9, pp. 686--699, May 2010.

\bibitem{Mudry.etal2012}
E.~Mudry, P.~C. Chaumet, K.~Belkebir, and A.~Sentenac, ``Electromagnetic wave
  imaging of three-dimensional targets using a hybrid iterative inversion
  method,'' \emph{Inv. Probl.}, vol.~28, no.~6, p. 065007, April 2012.

\bibitem{Jin.etal2017}
D.~Jin, R.~Zhou, Z.~Yaqoob, and P.~So, ``Tomographic phase microscopy:
  {P}rinciples and applications in bioimaging,'' \emph{J. Opt. Soc. Am. B},
  vol.~34, no.~5, pp. B64--B77, 2017.

\bibitem{Chaumet.Belkebir}
P.~C. Chaumet and K.~Belkebir, ``Three-dimensional reconstruction from real
  data using a conjugate gradient-coupled dipole method,'' \emph{Inv. Probl.},
  vol.~25, no.~2, p. 024003, 2009.

\bibitem{Belkebir.etal2005}
K.~Belkebir, P.~C. Chaumet, and A.~Sentenac, ``Superresolution in total
  internal reflection tomography,'' \emph{J. Opt. Soc. Am. A}, vol.~22, no.~9,
  pp. 1889--1897, September 2005.

\bibitem{Dubois.etal2005}
A.~Dubois, K.~Belkebir, and M.~Saillard, ``Retrieval of inhomogeneous targets
  from experimental frequency diversity data,'' \emph{Inv. Probl.}, vol.~21,
  no.~6, pp. S65--S81, 2005.

\bibitem{Tian.Waller2015}
L.~Tian and L.~Waller, ``{3D} intensity and phase imaging from light field
  measurements in an {LED} array microscope,'' \emph{Optica}, vol.~2, pp.
  104--111, 2015.

\bibitem{Kamilov.etal2015}
U.~S. Kamilov, I.~N. Papadopoulos, M.~H. Shoreh, A.~Goy, C.~Vonesch, M.~Unser,
  and D.~Psaltis, ``Learning approach to optical tomography,'' \emph{Optica},
  vol.~2, no.~6, pp. 517--522, June 2015.

\bibitem{Kamilov.etal2016}
------, ``Optical tomographic image reconstruction based on beam propagation
  and sparse regularization,'' \emph{IEEE Trans. Comp. Imag.}, vol.~2, no.~1,
  pp. 59--70,, March 2016.

\bibitem{Kamilov.etal2016a}
U.~S. Kamilov, D.~Liu, H.~Mansour, and P.~T. Boufounos, ``A recursive {B}orn
  approach to nonlinear inverse scattering,'' \emph{IEEE Signal Process.
  Lett.}, vol.~23, no.~8, pp. 1052--1056, August 2016.

\bibitem{Zhang.etal2016}
T.~Zhang, C.~Godavarthi, P.~C. Chaumet, G.~Maire, H.~Giovannini, A.~Talneau,
  M.~Allain, K.~Belkebir, and A.~Sentenac, ``Far-field diffraction microscopy
  at $\lambda/10$ resolution,'' \emph{Optica}, vol.~3, no.~6, pp. 609--612,
  June 2016.

\bibitem{Charnotskii2015}
M.~Charnotskii, ``Extended {H}uygens-{F}resnel principle and optical waves
  propagation in turbulence: discussion,'' \emph{J. Opt. Soc. Am. A}, vol.~32,
  no.~7, pp. 1357--1365, July 2015.

\bibitem{Liu.etal2015a}
H.-Y. Liu, E.~Jonas, L.~Tian, J.~Zhong, B.~Recht, and L.~Waller, ``{3D} imaging
  in volumetric scattering media using phase-space measurements,'' \emph{Opt.
  Express}, vol.~23, no.~11, pp. 14\,461--14\,471, June 2015.

\bibitem{Singh.etal2017}
A.~K. Singh, D.~N. Naik, G.~Pedrini, M.~Takeda, and W.~Osten, ``Exploiting
  scattering media for exploring {3D} objects,'' \emph{Light Sci. Appl.},
  vol.~6, no.~2, pp. 1--7, February 2017.

\bibitem{Liu.etal2017}
H.-Y. Liu, U.~S. Kamilov, D.~Liu, H.~Mansour, and P.~T. Boufounos,
  ``Compressive imaging with iterative forward models,'' in \emph{Proc. {IEEE}
  Int. Conf. Acoustics, Speech and Signal Process. ({ICASSP 2017})}, New
  Orleans, LA, USA, March 5-9, 2017, arXiv:1610.01852 [cs.CV].

\bibitem{Abubakar.etal2005}
A.~Abubakar, P.~M. {van den Berg}, and T.~M. Habashy, ``Application of the
  multiplicative regularized contrast source inversion method tm- and
  te-polarized experimental fresnel data,'' \emph{Inv. Probl.}, vol.~21, no.~6,
  pp. S5--S14, 2005.

\bibitem{Baussard2005}
A.~Baussard, ``Inversion of multi-frequency experimental data using an adaptive
  multiscale approach,'' \emph{Inv. Probl.}, vol.~21, no.~6, pp. S15--S32,
  2005.

\bibitem{Crocco.etal2005}
L.~Crocco, M.~{D'Urso}, and T.~Isernia, ``Testing the contrast source extended
  born inversion method against real data: the tm case,'' \emph{Inv}, vol.~21,
  no.~6, pp. S33--S51, 2005.

\bibitem{Feron.etal2005}
O.~F{\'e}ron, B.~Duch{\^e}ne, and A.~{Mohammad-Djafari}, ``Microwave imaging of
  inhomogeneous objects made of a finite number of dielectric and conductive
  materials from experimental data,'' \emph{Inv. Probl.}, vol.~21, no.~6, pp.
  S95--S117, 2005.

\bibitem{Born.Wolf2003}
M.~Born and E.~Wolf, \emph{Principles of Optics}, 7th~ed.\hskip 1em plus 0.5em
  minus 0.4em\relax Cambridge Univ. Press, 2003, ch. Scattering from
  inhomogeneous media, pp. 695--734.

\bibitem{Beck.Teboulle2009}
A.~Beck and M.~Teboulle, ``A fast iterative shrinkage-thresholding algorithm
  for linear inverse problems,'' \emph{SIAM J. Imaging Sciences}, vol.~2,
  no.~1, pp. 183--202, 2009.

\bibitem{Bishop1995}
C.~M. Bishop, \emph{Neural Networks for Pattern Recognition}.\hskip 1em plus
  0.5em minus 0.4em\relax Oxford, 1995.

\bibitem{Figueiredo.Nowak2003}
M.~A.~T. Figueiredo and R.~D. Nowak, ``An {EM} algorithm for wavelet-based
  image restoration,'' \emph{IEEE Trans. Image Process.}, vol.~12, no.~8, pp.
  906--916, August 2003.

\bibitem{Daubechies.etal2004}
I.~Daubechies, M.~Defrise, and C.~D. Mol, ``An iterative thresholding algorithm
  for linear inverse problems with a sparsity constraint,'' \emph{Commun. Pure
  Appl. Math.}, vol.~57, no.~11, pp. 1413--1457, November 2004.

\bibitem{Bect.etal2004}
J.~Bect, L.~Blanc-Feraud, G.~Aubert, and A.~Chambolle, ``A $\ell_1$-unified
  variational framework for image restoration,'' in \emph{Proc. {ECCV}},
  Springer, Ed., vol. 3024, New York, 2004, pp. 1--13.

\bibitem{LeCun.etal2015}
Y.~LeCun, Y.~Bengio, and G.~Hinton, ``Deep learning,'' \emph{Nature}, vol. 521,
  pp. 436--444, May 28, 2015.

\bibitem{Bottou2012}
L.~Bottou, \emph{Neural Networks: Tricks of the Trade}, 2nd~ed.\hskip 1em plus
  0.5em minus 0.4em\relax Springer, September 2012, ch. Stochastic Gradient
  Descent Tricks, pp. 421--437.

\bibitem{Mairal.etal2014}
J.~Mairal, F.~Bach, and J.~Ponce, ``Sparse modeling for image and vision
  processing,'' \emph{Foundations and Trends in Machine Learning}, vol.~8, no.
  2-3, pp. 1--199, 2014.

\bibitem{Bottou.etal2016}
L.~Bottou, F.~E. Curtis, and J.~Nocedal, ``Optimization methods for large-scale
  machine learning,'' June 2016, arXiv:1606.04838 [stat.ML].

\bibitem{Taflove.Hagness2005}
A.~Taflove and S.~C. Hagness, \emph{Computational {E}lectrodynamics: The
  {F}inite-{D}ifference {T}ime-{D}omain Method}, 3rd~ed.\hskip 1em plus 0.5em
  minus 0.4em\relax Artech House Publishers, 2005.

\bibitem{Geffrin.Sabouroux2009}
J.-M. Geffrin and P.~Sabouroux, ``Continuing with the {F}resnel database:
  experimental setup and improvements in {3D} scattering measurements,''
  \emph{Inv. Probl.}, vol.~25, no.~2, p. 024001, 2009.

\bibitem{Jackson1999}
J.~D. Jackson, \emph{Classical Electrodynamics}, 3rd~ed.\hskip 1em plus 0.5em
  minus 0.4em\relax Wiley, 1999.

\bibitem{Sheppard.Cogswell1990}
C.~Sheppard and C.~J. Cogswell, ``Three-dimensional image formation in confocal
  microscopy,'' \emph{J. Microsc.}, vol. 159, no.~2, pp. 179--194, August 1990.

\end{thebibliography}


\end{document}